\documentclass[preprint,12pt,authoryear]{elsarticle}

\journal{Remote Sensing of Environment}
\biboptions{authoryear,round}

\usepackage{amsmath,amssymb,amsfonts}
\usepackage{graphicx}
\usepackage{xcolor}
\usepackage{booktabs}
\usepackage{multirow}
\usepackage{tabularx}
\usepackage{array}
\usepackage{url}
\usepackage[hidelinks,hypertexnames=false]{hyperref}
\usepackage{algorithm}
\usepackage{placeins}
\usepackage{algpseudocode}
\usepackage{makecell}
\usepackage{lmodern}   
\usepackage{microtype} 
\usepackage{etoolbox}
\makeatletter
\patchcmd{\pprintMaketitle}{\footnotesize\itshape\elsaddress}%
  {\fontsize{7.6}{9.2}\selectfont\itshape\elsaddress}{}{}
\patchcmd{\MaketitleBox}{\footnotesize\itshape\elsaddress}%
  {\fontsize{7.6}{9.2}\selectfont\itshape\elsaddress}{}{}
\makeatother

\makeatletter
\newcommand*{\addFileDependency}[1]{
  \typeout{(#1)}
  \@addtofilelist{#1}
  \IfFileExists{#1}{}{\typeout{No file #1.}}
}
\makeatother

\newcolumntype{L}{>{\raggedright\arraybackslash}X}
\newcommand{\chfour}{CH$_4$}

\begin{document}

\begin{frontmatter}

\title{PlumeQuant:~Uncertainty-aware consistency assessment of methane plume masks and\\
emission-rate estimates}

\author[oudata,oudsai]{Parisa Masnadi Khiabani\corref{cor1}}
\ead{parisa.masnadi@ou.edu}

\author[uarizona2]{Wolfgang Jentner}
\author[oudsai,ouise]{Alireza Rangrazjeddi}
\author[oudata,ougis]{Michael C. Wimberly}
\author[oudata,oueec]{Binbin Weng}
\author[uarizona,uarizona2]{David Ebert}
\author[oudsai,ouise]{Charles Nicholson}

\cortext[cor1]{Corresponding author.}

\address[oudata]{Data Institute for Societal Challenges, University of Oklahoma, Norman, OK 73019, USA}
\address[oudsai]{Data Science \& Analytics Institute, University of Oklahoma, Norman, OK 73019, USA}
\address[oueec]{School of Electrical \& Computer Engineering, University of Oklahoma, Norman, OK 73019, USA}
\address[ouise]{School of Industrial \& Systems Engineering, University of Oklahoma, Norman, OK 73019, USA}
\address[ougis]{Department of Geography \& Environmental Sustainability, University of Oklahoma, Norman, OK 73019, USA}
\address[uarizona]{Department of Electrical \& Computer Engineering, University of Arizona, Tucson, AZ 85721, USA}
\address[uarizona2]{Office of Responsible AI, University of Arizona, Tucson, AZ 85721, USA}

\begin{abstract}
Imaging spectrometers increasingly distribute source-resolved methane plume products in which the plume mask, integrated mass enhancement (IME), plume length, emission rate, and uncertainty are physically and algorithmically linked. Using 63 EMIT-derived Carbon Mapper plume records from 27 scenes, we show that these published scalar quantities do not uniquely constrain the plume boundary: substantially different yet plausible masks reproduce the same IME, plume length, and emission rate. Genetic-algorithm (GA) ensembles conditioned on the published IME and plume length make this equifinality explicit: the high-confidence core selected by nearly all target-consistent masks covers a median of 13\% of the plausible footprint envelope, and ambiguity is largest for weak, low-overlap plumes. The diagnostics come from PlumeQuant, which recomputes IME, plume length, emission rate, and five-term uncertainty from distributed product components under stated conventions and evaluates four mask representations: the distributed reference mask, a transparent Carbon Mapper-informed analogue (CM-like), the GA ensemble, and optional expert edits. The CM-like mask is generated per plume without access to the reference mask or published quantities, with settings fixed once on a scene-disjoint 44-plume development split. It reproduced published IME with +0.72\% median difference and emission rate with +0.16\% (6.98\% mean absolute), reached 0.843 median intersection-over-union against the reference masks, and matched the published uncertainty scale (median ratio 1.01). Holdout mean absolute errors were 7.6\% (IME), 9.5\% (length), and 6.1\% (rate). These are product-level consistency diagnostics, not independent validation. They flag weak, offset, or ambiguous plumes for expert review.
\end{abstract}


\end{frontmatter}

\section{Introduction}
\label{sec:introduction}

Methane point-source monitoring has advanced rapidly with airborne and spaceborne imaging spectrometers that map localized \chfour{} column enhancements and support attribution to individual facilities or source regions \citep{frankenberg2016airborne,duren2019california,jacob2022quantifying,thorpe2023attribution}. 
This capability matters because a small number of strong point sources can contribute disproportionately to regional methane emissions and may be actionable once detected and localized \citep{duren2019california,cusworth2021intermittency}. 
The Earth Surface Mineral Dust Source Investigation (EMIT) provides a useful spaceborne case study: the instrument has been shown to attribute individual methane and carbon dioxide emission sources from space, and the EMIT methane plume-complex product provides plume-scale observations derived from imaging-spectrometer retrievals \citep{thorpe2023attribution,green2023emit}. 
We focus on EMIT-derived methane plume products distributed through the Carbon Mapper data system. Although the workflow is formulated for imaging-spectrometer plume products more broadly, all quantitative experiments reported here use EMIT plume scenes.

These products are increasingly used not as standalone images but as quantitative inputs to consequential decisions such as monitoring, reporting, and verification (MRV), regulatory and voluntary emissions reporting, facility-level attribution, and measurement-based emission inventories \citep{duren2019california,cusworth2021intermittency,jacob2022quantifying}. In each of these uses it is the reported emission rate and its stated uncertainty that are acted upon, even though that rate is the end of a processing chain running from the methane-enhancement raster through the plume mask, integrated mass enhancement (IME), and plume length to the wind speed. As these archives expand far faster than individual plumes can be checked against reference measurements, trustworthy use depends on transparent, reproducible ways to assess whether an individual plume product is internally consistent and to flag records whose reported quantities warrant closer review.

A plume product is not a single scalar estimate but a coupled set of components: a concentration raster, source location, plume mask, wind information, IME, plume length, an emission-rate estimate, and associated uncertainty. In what follows, we refer to IME, plume length, and emission rate collectively as the \emph{plume quantities}.
Carbon Mapper product documentation likewise describes plume products in terms of georeferenced plume imagery, IME, plume length, emission-rate information, wind information, quality flags, and uncertainty fields \citep{mapper2025product}. 
These quantities are physically and algorithmically linked: the plume mask determines which pixels contribute to IME, plume length sets the spatial scale used in the IME-to-rate conversion, and wind speed converts the observed excess mass into an emission rate \citep{frankenberg2016airborne,varon2018quantifying,thorpe2021improved}. 
As a result, differences in plume boundary, source position, wind field, concentration-to-mass conversion, or length convention propagate directly into every recomputed plume quantity.

Plume-mask delineation is therefore not merely a visualization step but a major source of uncertainty and a key driver of sensitivity in IME-based emission estimates. In imaging-spectrometer retrievals, plume boundaries are often hard to define because turbulence, source-location offsets, surface or instrument artifacts, and nearby sources distort the apparent plume structure. Recent EMIT work has further shown that orbital imaging-spectrometer plume catalogs can include marginal or false detections that require spectral and spatial vetting \citep{xiang2025identification}. For product users, the practical question after a plume is reported is not whether a new algorithm should replace the operational product, but whether the distributed components support a stable and internally consistent quantitative interpretation.

This product-level question is motivated by, but distinct from, prior work on methane remote-sensing uncertainty. Controlled-release studies have quantified detection probability and source-rate uncertainty under known-release conditions \citep{conrad2023robust,thorpe2021improved}, and single-blind satellite experiments have compared reported estimates against metered releases \citep{sherwin2023single,sherwin2024single}. These studies characterize technology-level performance. They do not address the question faced by a user of an individual plume product: do the distributed raster, mask, wind data, and published quantities support a mutually consistent interpretation? Answering it requires a workflow that recomputes plume quantities from the distributed components, evaluates mask sensitivity, and distinguishes convention-driven from physically meaningful disagreement.

Recent methodological work has concentrated on upstream detection and quantification rather than downstream consistency assessment of published plume products. Deep-learning approaches such as MethaNet estimate source rates directly from plume imagery \citep{jongaramrungruang2022methanet}, S2MetNet benchmarks methane point-source quantification from Sentinel-2 imagery \citep{radman2023s2metnet}, and CELNet targets plume identification from remotely sensed concentration images \citep{chen2025celnet}, and automated detection-and-monitoring pipelines now track methane super-emitters from satellite data at scale \citep{schuit2023automated}. Others have improved quantification from imaging spectroscopy or AI-based rate estimation \citep{pei2023improving,plewa2025improvements}, compared emerging hyperspectral satellite systems \citep{li2026comparing}, and reviewed the broader landscape of satellite methane point-source monitoring \citep{mohammadimanesh2025advancements}.
These efforts are complementary to PlumeQuant, which is designed to assess the internal consistency of EMIT methane plume products. PlumeQuant differs from 
end-to-end retrieval and quantification frameworks in a fundamental way: rather than producing new emission estimates from radiance or enhancement imagery, 
it starts from already distributed plume products and evaluates whether the raster, mask, source location, wind, plume-quantity, and uncertainty fields are mutually
consistent under explicit product conventions.

PlumeQuant operates downstream of detection and initial quantification as a consistency-checking layer. Starting from the distributed components (enhancement raster, source metadata, wind information, published quantities, and reference mask), it recomputes the plume quantities and their uncertainty under explicit conventions. The operational Carbon Mapper masking code is not used. Instead, PlumeQuant implements a transparent, description-guided \emph{dynamic-threshold mask}, referred to as the CM-like mask (Section~\ref{sec:cm_like_mask}). This mask is a reproducible baseline for testing how mask choice affects the recomputed quantities, not a new segmentation algorithm or an official reproduction of the Carbon Mapper approach. PlumeQuant additionally uses a constrained genetic algorithm (GA) as an inverse, target-conditioned mask search: given the published IME and plume length, it asks which spatially plausible masks satisfy those quantities. The answer reveals whether the published scalars correspond to a unique boundary or admit substantially different plausible ones. For every mask, PlumeQuant recomputes plume quantities and uncertainty and evaluates both scalar and spatial agreement. 

The contributions of this study are fourfold. First, we introduce a consistency assessment framework that evaluates agreement between published EMIT plume metadata and raster-based recomputations derived from CM-like, GA-candidate, and expert-edited masks. Second, we characterize plume-mask sensitivity using a transparent dynamic-threshold baseline together with a constrained GA search that explores plausible masks consistent with the published IME and length, while explicitly avoiding interpretation of the GA result as ground truth. Third, we propagate emission-rate uncertainty from wind speed, retrieval (concentration) uncertainty, mask selection, plume length, and concentration-to-mass conversion, and we report both the combined uncertainty and the dominant component contributions. Fourth, we provide a review workflow that supports single-plume inspection, batch consistency assessment, and provenance tracking. The released tool additionally allows an analyst to adjust a plume boundary, guided by the footprint-confidence map, and to recompute all plume quantities and uncertainty terms through the same conventions used for the automated masks. Expert edits are used in this study only for representative-case interpretation, not for benchmark-level statistics.

\section{Data and methods}
 
\subsection{EMIT-derived Carbon Mapper methane plume products}
\label{sec:data_products}
This study uses publicly available Carbon Mapper methane plume products generated from EMIT imaging-spectrometer observations \citep{thorpe2023attribution,mapper2025product}. EMIT is a NASA Jet Propulsion Laboratory imaging spectrometer installed on the International Space Station that measures reflected solar radiation across the visible to shortwave-infrared spectral range \citep{NASAEMITInstrumentSpecs}. The Carbon Mapper data system distributes plume products from multiple satellite and airborne hyperspectral sensors, among them the products generated from EMIT observations \citep{mapper2025product}. For the EMIT-based products considered here, the Carbon Mapper Product Guide reports an approximate ground sampling distance of 50--60~m, spectral coverage from 381 to 2493~nm, and spectral spacing of approximately 7.5~nm \citep{mapper2025product}. These sensor and product characteristics define the spatial and spectral context for the plume-mask, plume-length, and emission-rate consistency analysis.
 
\begin{figure*}
\centering
\includegraphics[width=0.99\textwidth]{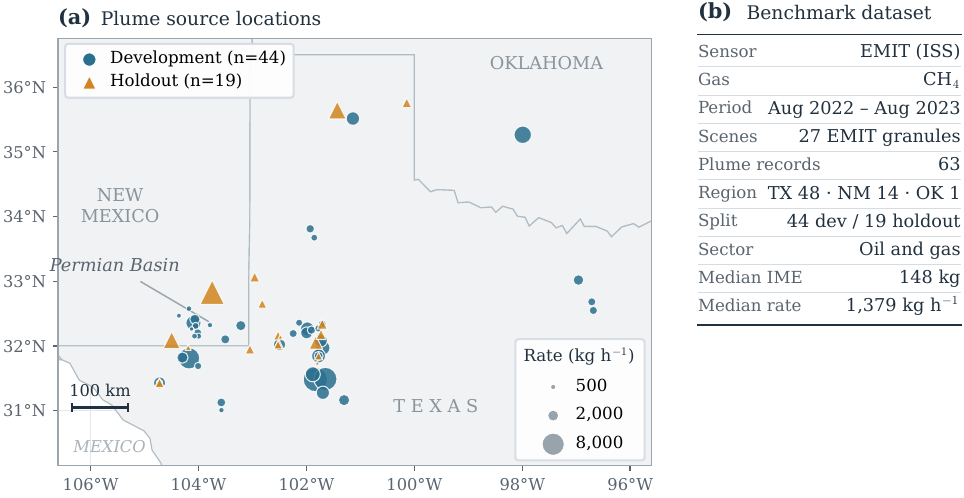}
\caption{Benchmark dataset overview. (a) Geographic distribution of the 63 EMIT/Carbon Mapper methane plume records from 27 scenes in the Texas--New Mexico--Oklahoma study area, which spans the Permian Basin and adjacent regions. Marker area is proportional to the published emission rate, and marker color/shape indicates the fixed, scene-disjoint development ($n=44$) and holdout ($n=19$) splits. 
(b) Summary of the benchmark: sensor/platform, gas, acquisition period, regional distribution, dominant source sector, median published IME, median published emission rate, and development/holdout partition. 
State boundaries are drawn from a generalized public-domain dataset, are approximate, and are shown for geographic orientation only. They do not imply any position concerning the delimitation of borders.}
 \label{fig:benchmark_map}
 \end{figure*}

 
Public product components were harmonized into a plume-level analysis inventory for this study. The inventory links each Carbon Mapper plume record to source coordinates, acquisition metadata, product-version information, methane-enhancement imagery, available plume-mask data, wind information, and published plume-level quantities. This inventory is an organizational layer developed for reproducible analysis. It is not an additional Carbon Mapper product. Table~\ref{tab:input_data_components} summarizes the harmonized data components, their file formats, and their role in the PlumeQuant consistency analysis.
\begin{table}[htbp]
\centering
\caption[Input data components used by PlumeQuant]{Input data components used by PlumeQuant. The table summarizes the data products harmonized for each plume record.}
\label{tab:input_data_components}
\scriptsize
\setlength{\tabcolsep}{3pt}
\renewcommand{\arraystretch}{1.15}
 
\begin{tabularx}{\linewidth}{@{}
>{\raggedright\arraybackslash}p{0.15\linewidth}
>{\raggedright\arraybackslash}p{0.21\linewidth}
>{\raggedright\arraybackslash}X
>{\raggedright\arraybackslash}X
@{}}
\toprule
\textbf{Data component} &
\textbf{Example file/source and format} &
\textbf{Key contents} &
\textbf{Role in PlumeQuant} \\
\midrule
 
Plume manifest &
\texttt{plume\_manifest.csv}\par
\emph{CSV table} &
Plume ID, source latitude/longitude, acquisition time, gas, instrument,
published IME, plume length, emission rate, emission-rate uncertainty, wind
speed/direction, quality flags, product-version fields, and raster paths &
Links each plume to its raster products and provides the published metadata
used for consistency comparisons \\
 
\addlinespace[2pt]
 
Scene methane column-enhancement raster &
\texttt{cmf.tif}\par
\emph{Single-band GeoTIFF} &
Scene-level methane column-enhancement or matched-filter values, coordinate
reference system, affine transform, pixel size, and nodata value &
Primary raster used to generate CM-like and GA candidate masks and to recompute
mask-based IME \\
 
\addlinespace[2pt]
 
Plume concentration product &
\texttt{ime-cmf-}\par
\texttt{concentrations.tif}\par
\emph{Single-band GeoTIFF} &
Plume-level concentration image distributed with the product. Finite pixels
define the delivered plume concentration crop &
Provides product-scale concentration context and supports reference-product
comparison \\
 
\addlinespace[2pt]
 
Reference plume mask &
\texttt{ime-cmf-mask.tif}\par
\emph{Binary/integer GeoTIFF} &
Product-delivered plume delineation, with active plume pixels separated from
background or nodata pixels &
Used as the spatial reference mask for IoU/Dice comparison and for reference-mask
recomputation of IME, plume length, and rate \\
 
\addlinespace[2pt]
 
Retrieval-uncertainty raster &
\texttt{uncertainty.tif}\par
\emph{Single-band GeoTIFF} &
Per-pixel uncertainty associated with the scene column-enhancement raster &
Supplies the retrieval component of IME and emission-rate uncertainty and
supports uncertainty-coverage checks \\
 
\addlinespace[2pt]
 
Artifact mask &
\texttt{artifact-mask.tif}\par
\emph{Single-band GeoTIFF} &
Scene artifact or quality flags on the raster grid &
Used to report how much of a candidate plume mask overlaps flagged pixels \\
 
\addlinespace[2pt]
 
HRRR meteorology &
NOAA HRRR cached subsets\par
\emph{GRIB2 / derived table} &
10~m wind components, 2~m temperature, surface pressure, specific humidity,
boundary-layer height, and model orography sampled near the plume time/location &
Provides wind speed/direction and environmental terms used in emission-rate
recomputation, uncertainty propagation, and concentration-to-mass factor
diagnostics \\
 
\bottomrule
\end{tabularx}
\end{table}
The study area and sample size follow from a census-style acquisition rather than a curated selection. The project's initial focus was Oklahoma and its oil and gas sector. Because EMIT plume detections over Oklahoma itself were sparse during the study period, the search domain was defined as Oklahoma together with the adjacent oil-and-gas-producing region of Texas and New Mexico, dominated by the Permian Basin. Within this Texas--New Mexico--Oklahoma domain, all EMIT-derived Carbon Mapper methane plume records available at the time of data retrieval for the August 2022--August 2023 acquisition period were downloaded. The analyzed sample comprises the resulting 63 plume records from 27 unique EMIT scenes (Figure~\ref{fig:benchmark_map}).
Records were included when the methane-enhancement raster, plume source location, and published plume quantities required for IME, plume length, and emission-rate comparison were available.
All 63 included records carried a distributed reference mask, so spatial mask comparisons (IoU, Dice) use the full \(n=63\) benchmark. One record lacked a published 
emission rate and its associated rate uncertainty. This record is retained for IME, length, and mask comparisons but is excluded from the emission-rate and uncertainty 
comparisons, which therefore use \(n=62\).
The distributed plume mask is therefore used as the product reference for spatial comparison and reference-mask recomputation, rather than as an independent ground truth.
For convention and parameter selection, the 63-plume benchmark is partitioned once at the scene level into a development set (44 plumes) and a scene-disjoint holdout set (19 plumes). All calculation conventions and mask-generation settings are selected on the development split alone. The holdout is then scored once, as an independent check, after the configuration is frozen. The assignment procedure and per-plume results are detailed in Section~\ref{sec:results} and Supplementary Section~S5.

\subsection{PlumeQuant workflow overview}
 
PlumeQuant is an uncertainty-aware consistency-assessment framework for publicly available Carbon Mapper methane plume products generated from EMIT observations. It operates downstream of Carbon Mapper product generation and uses the distributed product components to recompute the plume quantities and their uncertainty under explicit assumptions. This framing follows Carbon Mapper's quantification chain, in which detected enhancements are segmented into a plume mask and emission rates are estimated from IME, plume length, and surface wind speed \citep{duren2025carbon}, consistent with the IME--length--wind relationship of prior studies \citep{varon2018quantifying,duren2019california}.
 
\begin{figure}[htbp]
\centering
\includegraphics[width=0.97\textwidth]{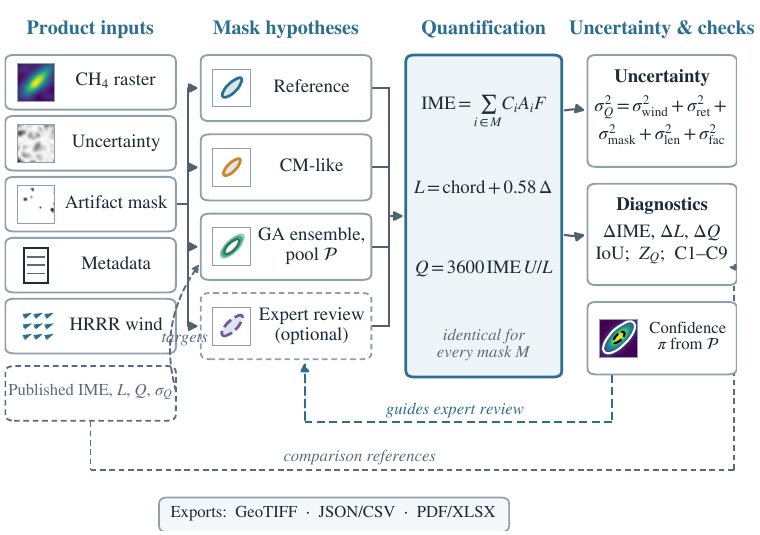}
\caption{PlumeQuant workflow for product-level consistency assessment. Product inputs (left) feed four mask hypotheses: the distributed reference mask, the CM-like dynamic-threshold mask, the GA ensemble with its pool \(\mathcal{P}\), and an optional expert-reviewed mask. Each mask passes through the same fixed quantification conventions to recompute IME, plume length, and emission rate, which feed the five-term uncertainty propagation and the consistency diagnostics (recomputed--published differences, IoU/Dice, \(Z_Q\), checks C1--C9; Supplementary Section~S7). Published quantities (dashed) enter only as GA search targets and comparison references. The footprint-confidence map guides the expert-review loop, and results are exported with full provenance.}
\label{fig:workflow_dashboard}
\end{figure}

Figure~\ref{fig:workflow_dashboard} summarizes the PlumeQuant workflow for a representative plume record. 
The workflow links harmonized product components to four alternative mask representations: the distributed reference mask, CM-like mask, GA candidate masks, and expert-edited masks when available.
For each mask, PlumeQuant recomputes the plume quantities and uncertainty, then evaluates scalar agreement, spatial agreement, uncertainty contributions, and mask sensitivity within a reproducible review workflow.  It also records nine automated mutual-consistency checks (C1--C9) that compare the recomputed quantities against the published fields, distributed reference mask, HRRR meteorology, artifact mask, and uncertainty-component structure. These checks are defined in Supplementary Section~S7. Per-plume check outcomes are recorded in the archived analysis exports. In this paper the checks serve as review-flagging machinery rather than as headline results.
 
\subsection{Product harmonization and spatial domain}

Before quantification, all product components were organized at the plume level and checked for coordinate consistency. Source locations, methane-enhancement rasters, distributed product masks, and candidate masks were evaluated in georeferenced map coordinates so that plume area, plume length, and mask overlap could be computed in physical units. Pixel area was derived from the raster geotransform for each scene, and mask comparisons were performed on a common raster grid after spatial alignment when required. Detailed raster input rules, coordinate transformations, crop definitions, nodata treatment, and grid-alignment procedures are provided in Supplementary Section~S1.
\subsection{Meteorology and concentration-to-mass conversion}
\label{sec:meteorology_conversion}
Wind speed is required to convert IME and plume length into a diagnostic instantaneous emission rate. For each plume, PlumeQuant uses the wind information linked to the Carbon Mapper product and, when available, samples HRRR meteorological fields near the plume time and location. HRRR is a NOAA atmospheric model with hourly updates and 3~km grid spacing \citep{dowell2022high,noaa_hrrr}. Carbon Mapper also reports the use of HRRR wind fields for U.S. observations in its plume-quantification workflow \citep{carbonmapper2024atbd,duren2025carbon}.

When HRRR fields are available, PlumeQuant samples the analysis field closest to the plume acquisition time together with the adjacent hourly fields. 
Wind speed is derived from the HRRR 10~m wind components (\(u_{10},v_{10}\)). The closest hourly analysis and its two neighboring hours 
(\(\pm 1\)~h) are used to characterize short-term temporal variability.
Within each field, a local neighborhood of grid cells around the reported source is used to represent meteorological variability near the plume. The effective wind speed is calculated as the mean of the local sampled speeds, and wind-speed uncertainty is estimated from the standard deviation of all sampled speeds across the local neighborhood and the three hourly fields. Wind direction is summarized using circular statistics and is used only for diagnostics (for example, upwind-fraction penalties in the candidate-mask search). It does not modify the mask used for quantification. 
No empirical wind-speed calibration or correction factor is applied. When HRRR fields are unavailable for a record, PlumeQuant falls back to the wind information 
linked to the Carbon Mapper product. Neighborhood size and sampling settings are reported in Supplementary Section~S1.
This treatment represents HRRR wind as a plume-scale meteorological input rather than an exact in-plume velocity field.

Methane column enhancement must be converted to methane mass per unit area before IME can be calculated. A concentration-path enhancement, commonly expressed in ppm~m, represents an excess methane mole fraction integrated over path length. Under pressure \(P\) and temperature \(T\), the corresponding concentration-to-mass conversion factor is
\begin{equation}
F(P,T) = 10^{-6}\,\frac{M_{\mathrm{CH_4}}\,P}{R\,T},
\end{equation}
where \(F\) has units of \(\mathrm{kg\,m^{-2}}\) per \(\mathrm{ppm\,m}\), \(M_{\mathrm{CH_4}}\) is the molar mass of methane, and \(R\) is the universal gas constant. For a pixel with methane enhancement \(C_i\) and area \(A_i\), the corresponding methane mass is
\begin{equation}
m_i = C_i\,A_i\,F .
\end{equation}
For the product-convention-aligned recomputation used here, \(F\) is evaluated from the AFGL midlatitude summer standard atmosphere at the scene ground elevation. The AFGL profiles provide pressure and temperature as functions of altitude for a set of standard atmospheres \citep{anderson1986afgl}. Let \(P_{\mathrm{MLS}}(z)\) and \(T_{\mathrm{MLS}}(z)\) denote pressure and temperature from the midlatitude summer profile at ground elevation \(z\). The applied conversion factor is
\begin{equation}
F_{\mathrm{MLS}}(z) = 10^{-6}\,
\frac{M_{\mathrm{CH_4}}\,P_{\mathrm{MLS}}(z)}{R\,T_{\mathrm{MLS}}(z)} .
\end{equation}
This scalar convention is consistent with the Carbon Mapper L3/L4 quantification description, in which IME is computed from segmented plume pixels using a factor that converts \(\mathrm{ppm\,m}\) to \(\mathrm{kg\,m^{-2}}\) \citep{carbonmapper2024atbd,duren2025carbon}.
It also empirically reproduces the published IME values without plume-specific scaling (Section~\ref{sec:results_conversion}); 
we therefore treat it as a product-convention inference rather than a statement of Carbon Mapper's exact internal profile.
It is also consistent with prior imaging-spectrometer work showing that retrieved enhancement, IME, and flux estimates depend on atmospheric and geometric assumptions in the enhancement spectrum \citep{foote2021impact}.

Differences among concentration-to-mass conversion conventions affect both IME and emission rate because the conversion factor multiplies every selected plume pixel. We therefore treat \(F\) as a product convention and apply the same value of \(F_{\mathrm{MLS}}(z)\) to all masks for a given plume. To test whether this scalar standard-atmosphere convention was consistent with the published IME values, we also evaluated predefined pressure--temperature alternatives: HRRR boundary-layer, local-surface, planetary-boundary-layer (PBL)-mean, U.S. Standard Atmosphere 1976, and fixed normal-temperature-and-pressure conventions. These alternatives were used only for convention-sensitivity analysis. The uncertainty associated with \(F\) is treated separately in the uncertainty propagation section.
\subsection{Plume quantification and reference mask recomputation}
\label{sec:plume_quantification}
 
For each plume mask \(M\), PlumeQuant recomputes the plume quantities using the same methane-enhancement raster, meteorological input, and concentration-to-mass conversion convention. This fixed-quantification setup allows differences among the distributed reference mask, CM-like mask, GA candidate mask, and any expert-edited mask to be interpreted as mask sensitivity. The recomputation follows the IME-based plume-quantification framework used in prior methane point-source studies \citep{frankenberg2016airborne,varon2018quantifying,duren2019california,duren2025carbon}.
The IME for a mask \(M\) is calculated as
\begin{equation}
\mathrm{IME}(M)
=
\sum_{i \in M}
C_i A_i F ,
\label{eq:ime}
\end{equation}
where \(C_i\) is the methane column enhancement for pixel \(i\), \(A_i\) is the pixel area, and \(F\) is the concentration-to-mass conversion factor. The summation includes only finite enhancement values inside the selected mask. For a given plume, the same value of \(F\) is used for all mask representations, so differences in recomputed IME reflect differences in the selected pixels rather than differences in the conversion convention.
 
Plume length \(L(M)\) represents the horizontal spatial scale used to convert IME into an emission rate. Because the emission rate is inversely proportional to plume length, small differences in how length is measured from a pixelated mask can lead to systematic differences in the recomputed rate. We therefore define plume length as a fixed calculation convention and apply the same convention to all masks.
 
\begin{equation}
L(M)
=
\max_{i,j \in M}
\left\| \mathbf{x}_i - \mathbf{x}_j \right\|
+
0.58\,\Delta ,
\label{eq:length_operator}
\end{equation}
where \(\mathbf{x}_i\) and \(\mathbf{x}_j\) are the map coordinates of selected pixel centers, and \(\Delta\) is the pixel spacing. The \(0.58\,\Delta\) correction accounts for the fact that a plume can extend beyond the centers of the endpoint pixels. This correction was selected once using the development plumes to match the published plume-length convention and was then held fixed for the holdout evaluation. The same length operator was applied to the distributed reference mask, the CM-like mask, the GA candidate mask, and any expert-edited mask. Alternative length definitions are reported only as length-convention sensitivity tests.

The diagnostic emission rate for mask \(M\) is 
\begin{equation}
Q(M)
=
3600
\frac{\mathrm{IME}(M) U}
{L(M)} ,
\label{eq:rate}
\end{equation}
where \(Q(M)\) is in kg~h\(^{-1}\), \(U\) is wind speed in m~s\(^{-1}\), and \(L(M)\) is plume length in m. The factor 3600 converts seconds to hours. This equation follows the IME-based relation between plume mass, plume length, wind speed, and source rate used in methane plume quantification studies~\citep{varon2018quantifying,duren2025carbon}.
 
Reference-mask recomputation applies Eqs.~\ref{eq:ime}--\ref{eq:rate} to the distributed Carbon Mapper mask and provides a product-consistency baseline: if the components are consistent under the stated assumptions, the recomputed quantities should match the published values within the expected uncertainty range. Alternative masks then pass through the same equations, so differences among masks are interpreted as mask sensitivity. The distributed mask is a product reference for comparison, not independent ground truth.
\subsection{Carbon Mapper-informed dynamic-threshold mask} \label{sec:cm_like_mask}
The CM-like mask, previewed in the Introduction, is a transparent, description-guided \emph{dynamic-threshold} baseline for plume-mask sensitivity analysis. Carbon Mapper describes plume segmentation as the step that separates background pixels from enhanced methane or carbon dioxide pixels to form a plume boundary for mass and emission quantification \citep{carbonmapper2024atbd,duren2025carbon}. Carbon Mapper's operational masking code is not publicly available, so the implementation used here does not reproduce it. Instead, it is a description-guided analogue that uses source-centered thresholding, spatial component selection, and deterministic post-processing to produce a reproducible candidate mask for consistency testing.

The source-centered threshold is computed as
\begin{equation}
T = \frac{1}{K^{\ast}} \sum_{k=1}^{K^{\ast}} t_k ,
\label{eq:cm_threshold}
\end{equation}
where \(t_k\) is the enhancement threshold obtained from the \(k\)-th predefined distance--percentile ring, i.e., the specified percentile of finite enhancement values within that source-centered ring (Supplementary Section~S2), and \(K^{\ast}\) is the number of valid radius--percentile thresholds. Pixels with finite methane enhancement greater than \(T\) form the initial binary mask.

The initial binary mask is filtered using deterministic spatial criteria. Small isolated components are removed, and components far from the reported source are excluded. When one remaining component contains the source pixel, that component is selected. If the source pixel is not included in any component, the method selects the component with the strongest enhancement after applying a mild source-distance penalty. If another known source lies inside the selected component, watershed separation can be used to divide the component into source-associated regions, and only the region associated with the target source is retained \citep{vincent1991watersheds}. Small internal holes are filled, and the source pixel is added only when it remains connected to the selected plume body.

The resulting CM-like mask is used only as a reproducible baseline for sensitivity analysis. Its role is not to reproduce the operational Carbon Mapper mask, but to provide a transparent mask generated from the same methane-enhancement raster and source information. The numerical settings (radius--percentile pairs, component-size filters, source-distance limits, and post-processing thresholds) were selected using only the development split and then fixed before holdout evaluation. Specifically, the error-tuned constants (a background-clip threshold defined by a median-absolute-deviation multiplier with lower and upper band margins, the internal hole-filling limit, and an optional downwind trim) were chosen by a grid search over the development plumes that minimized the mean absolute emission-rate difference relative to the published rates, subject to two non-regression constraints on spatial agreement with the distributed reference masks: no protected development plume could lose more than 0.05 IoU relative to the component-selection-only baseline, and the development median IoU could not fall below that baseline. Component selection itself (the source-containing component, otherwise the component with the maximum integrated enhancement) is an architectural rule applied in every configuration and was not error-fitted. The optional downwind trim was rejected on the development split and is disabled in the frozen configuration. The frozen constants were then evaluated on the holdout exactly once. These implementation details are reported in Supplementary Section~S2.
\subsection{Genetic algorithm candidate masks and footprint confidence}
\label{sec:GA}

The genetic algorithm is used as a target-conditioned candidate-mask search. Its purpose is not to independently detect a plume, validate the true plume extent, or characterize the full uncertainty of all possible plume boundaries. Instead, given the published IME and plume length, the GA addresses a narrower diagnostic question: which spatially plausible masks can reproduce those published quantities under the same quantification assumptions? This inverse search is useful because agreement in scalar plume quantities does not necessarily imply a unique plume boundary. If multiple distinct masks reproduce nearly the same published IME and plume length, then those scalar quantities leave part of the plume geometry weakly constrained. The GA therefore provides a spatial diagnostic of mask-boundary non-uniqueness by identifying alternative target-consistent plume boundaries and showing which regions of the footprint are stable, weakly constrained, or ambiguous. Because the published IME and plume length are included in the search objective, agreement between a GA-derived mask and the published scalar quantities is interpreted as target consistency rather than independent validation of the plume boundary.

The GA is implemented as a single-objective evolutionary optimization using DEAP \citep{fortin2012deap}, following the general genetic-algorithm framework introduced for population-based search and selection \citep{holland1975adaptation}. Each GA individual represents a binary plume mask within a source-centered computational search domain constructed from finite methane-enhancement pixels around the reported source. This local domain reduces computational cost relative to searching the full scene raster while remaining large enough to contain the plume neighborhood considered in the consistency analysis. When available, a buffered version of the CM-like mask is used only to help define the local search domain and keep the optimization focused on the relevant plume area. It is not used as an objective target. For a candidate mask \(M\), the objective function is
\begin{equation}
\begin{split}
J(M) &= w_I E_I(M) + w_L E_L(M) + w_{\mathrm{low}} P_{\mathrm{low}}(M) \\
     &\quad + w_{\mathrm{comp}} P_{\mathrm{comp}}(M) + w_{\mathrm{src}} P_{\mathrm{src}}(M) \\
     &\quad + w_{\mathrm{shape}} P_{\mathrm{shape}}(M) + w_{\mathrm{up}} P_{\mathrm{up}}(M).
\end{split}
\label{eq:ga_objective}
\end{equation}
Here, \(E_I(M)\) and \(E_L(M)\) measure agreement with the published IME and published plume length, and the \(P\) terms are soft spatial-plausibility penalties. These penalties discourage masks that are dominated by weak-enhancement pixels, split into multiple disconnected components, drift far from the reported source, form highly irregular boundaries, or place a large fraction of selected pixels upwind of the source.

The IME and plume-length target errors are defined as
\begin{equation}
E_I(M)
=
\frac{\left| \mathrm{IME}(M) - \mathrm{IME}_{\mathrm{pub}} \right|}
     {\max\left( \left| \mathrm{IME}_{\mathrm{pub}} \right|, \epsilon \right)},
\qquad
E_L(M)
=
\frac{\left| L(M) - L_{\mathrm{pub}} \right|}
     {\max\left( \left| L_{\mathrm{pub}} \right|, \epsilon \right)} ,
\label{eq:ga_target_errors}
\end{equation}
where \(\mathrm{IME}_{\mathrm{pub}}\) and \(L_{\mathrm{pub}}\) are the published IME and plume length, and \(\epsilon\) is a small positive constant used to avoid division by zero. The spatial penalties are needed because the inverse search is highly underdetermined: without regularization the GA could satisfy the scalar targets with fragmented, off-source, weak-enhancement, or highly irregular masks that say nothing about plausible boundary variability. They are soft regularization terms, not physical laws, so the ensemble is a diagnostic sample of target-consistent, spatially plausible masks rather than uncertainty over all mathematically possible pixel combinations. The lowest-objective mask from the fixed GA configuration is reported as the GA candidate mask, \(M_{\mathrm{GA}}\). Population size, generation count, random seed, objective weights, and repair rules are documented in Supplementary Section~S3.

A single GA run returns one best candidate mask, but this mask alone does not indicate whether other plausible plume boundaries can reproduce the same published plume quantities. To examine target-consistent mask variability, the GA is run as an ensemble using \(R = 8\) independent random seeds. Across the ensemble, candidate masks are retained when their recomputed IME and plume length are both within predefined relative-error tolerances of the published values:
\begin{equation}
\mathcal{P}
=
\left\{
M :
\frac{|\mathrm{IME}(M)-\mathrm{IME}_{\mathrm{pub}}|}
{\max(|\mathrm{IME}_{\mathrm{pub}}|,\epsilon)}
\le \tau_I
\ \text{and}\
\frac{|L(M)-L_{\mathrm{pub}}|}
{\max(|L_{\mathrm{pub}}|,\epsilon)}
\le \tau_L
\right\},
\label{eq:equifinal_pool}
\end{equation}
where \(\tau_I = \tau_L = 0.15\). We refer to \(\mathcal{P}\) as the equifinal pool: a set of distinct masks that produce nearly equivalent published-scale quantities under the same quantification assumptions. The term is borrowed from the equifinality concept in environmental modelling, in which distinct model configurations reproduce the same observed behaviour \citep{beven2006manifesto}. If no masks satisfy these criteria, the plume is flagged as a failed target-consistency case and no footprint-confidence map is interpreted for that plume.

The tolerances set how closely pool members must reproduce the published quantities, so the footprint-confidence map represents boundary variability conditional on these tolerances, the GA search design, and the spatial regularization. Its purpose is to summarize variability among plume-like masks consistent with the published IME and length, not to recover all possible masks.

For each pixel \(i\) in the GA candidate region, the \emph{footprint confidence} is defined as the inclusion frequency across the equifinal pool:
\begin{equation}
\pi_i
=
\frac{1}{|\mathcal{P}|}
\sum_{M \in \mathcal{P}}
\mathbf{1}\!\left[\, i \in M \,\right],
\label{eq:footprint_confidence}
\end{equation}
where \(\mathbf{1}\!\left[\, i \in M \,\right]\) equals one when pixel \(i\) is included in mask \(M\) and zero otherwise. Pixels with \(\pi_i\) close to one are selected by nearly all equifinal masks, while pixels with \(\pi_i\) close to zero are rarely selected. Pixels with \(\pi_i > 0\) form the union envelope, and pixels with \(\pi_i \ge 0.9\) form the high-confidence core. Intermediate values identify an ambiguous fringe where different plausible masks make different boundary choices.

The footprint-confidence map is an exploratory diagnostic of mask non-uniqueness conditional on the targets and plausibility terms. It is not a probability map of the true boundary and is not included in the propagated rate uncertainty. A compact high-confidence footprint means the published scalars strongly constrain the geometry. A broad ambiguous fringe identifies plumes where boundary choice is weakly constrained and expert review is most informative.

Three outcomes are diagnostic: a GA mask that matches the published scalars but overlaps the reference mask poorly signals boundary non-uniqueness; agreement with both supports product-level consistency; and failure to reproduce the published quantities within the search constraints flags the record for review of its raster, source location, wind input, product version, or mask.
 
\subsection{Expert edited masks}
 
Expert-edited masks provide an optional review layer for cases requiring visual interpretation. An edited mask is an alternative hypothesis for the same plume, not ground truth. Editing is GA-informed rather than free-hand: the analyst is shown the footprint-confidence map (Eq.~\ref{eq:footprint_confidence}) with its union envelope and high-confidence core, may initialize from a member of the equifinal pool \(\mathcal{P}\), and may then add or remove pixels guided by the enhancement raster, source location, plume morphology, and visible artifacts. The edited mask \(M_{\mathrm{edit}}\) is recomputed with the same meteorological inputs, conversion factor, length convention, and rate equation as all other masks. Each edit is recorded with its provenance (seed mask, pixel changes, analyst note), and edited-mask results are reported separately so analyst intervention does not affect the automated-mask comparisons.
\subsection{Uncertainty propagation and sensitivity analysis}
\label{sec:uncertainty_methods}
 
Uncertainty and sensitivity are treated as related but distinct quantities in
this study. Uncertainty describes the expected spread in a recomputed emission
rate for a given mask and set of input assumptions. Sensitivity describes how
the plume quantities change when the plume mask or selected input
convention changes. This distinction is important because a plume can have a
well-defined recomputed rate for a selected mask while still showing strong
sensitivity to plausible mask boundary choices.
 
For each plume, mask sensitivity is evaluated by comparing recomputed quantities
across the available mask set,
\begin{equation}
\mathcal{M}
=
\{
M_{\mathrm{ref}},
M_{\mathrm{CM}},
M_{\mathrm{GA}},
M_{\mathrm{edit}}
\},
\end{equation}
where \(M_{\mathrm{ref}}\) is the product reference mask, \(M_{\mathrm{CM}}\) is
the Carbon Mapper-informed dynamic-threshold mask, \(M_{\mathrm{GA}}\) is the
genetic algorithm candidate mask, and \(M_{\mathrm{edit}}\) is an optional expert-edited mask. For any plume quantity \(X \in \{\mathrm{IME}, L, Q\}\), the
sensitivity of mask \(M_j\) relative to the reference mask is
\begin{equation}
S_X(M_j)
=
100
\frac{
X(M_j)-X(M_{\mathrm{ref}})
}
{
\max(|X(M_{\mathrm{ref}})|,\epsilon)
},
\label{eq:sensitivity_pairwise}
\end{equation}
where \(\epsilon\) is a small positive constant used only to avoid division by
zero. The across mask sensitivity range is
\begin{equation}
S_X^{\mathrm{range}}
=
100
\frac{
\max_{M_j \in \mathcal{M}} X(M_j)
-
\min_{M_j \in \mathcal{M}} X(M_j)
}
{
\max(|X(M_{\mathrm{ref}})|,\epsilon)
}.
\label{eq:sensitivity_range}
\end{equation}
Equations~\ref{eq:sensitivity_pairwise} and \ref{eq:sensitivity_range} provide
deterministic sensitivity diagnostics. They are reported separately from the
propagated rate uncertainty.
 
The propagated emission rate uncertainty is estimated with first-order error
propagation \citep{jcgm2008evaluation}; comparable structured uncertainty budgets have been developed for satellite methane point-source quantification \citep{gorrono2023understanding}. For the rate equation in
Eq.~\ref{eq:rate}, PlumeQuant represents the total uncertainty as the quadrature
combination of wind, retrieval, mask, plume length, and conversion factor terms,
 
\begin{equation}
\sigma_Q
=
\left[
\sigma_{Q,\mathrm{wind}}^2
+
\sigma_{Q,\mathrm{ret}}^2
+
\sigma_{Q,\mathrm{mask}}^2
+
\sigma_{Q,\mathrm{len}}^2
+
\sigma_{Q,\mathrm{fac}}^2
\right]^{1/2}.
\label{eq:total_uncertainty}
\end{equation}
Here, \(\sigma_{Q,\mathrm{wind}}\) represents wind-speed uncertainty, \(\sigma_{Q,\mathrm{ret}}\) represents retrieval or concentration uncertainty, \(\sigma_{Q,\mathrm{mask}}\) represents uncertainty associated with mask-boundary selection, \(\sigma_{Q,\mathrm{len}}\) represents plume-length uncertainty, and \(\sigma_{Q,\mathrm{fac}}\) represents uncertainty in the concentration-to-mass conversion factor. This equation provides a first-order uncertainty envelope for recomputed emission rates under the assumptions used in PlumeQuant.

Four of the five components follow first-order (Taylor) propagation, \(\sigma_{Q,k} = \lvert \partial Q/\partial x_k \rvert\,\sigma_{x_k}\): the magnitude of the partial derivative of the rate \(Q\) with respect to input \(x_k\) multiplied by that input's standard deviation. The terms are treated as mutually independent and combine in quadrature in Eq.~\ref{eq:total_uncertainty}. Because \(Q = 3600\,\mathrm{IME}\,U/L\) is proportional to IME and to \(U\) and inversely proportional to \(L\), the wind, plume-length, and conversion-factor terms reduce to the relative form \(|Q|\,\sigma_x/x\), while the retrieval term propagates the pixel-level concentration uncertainty through IME (below). The mask term is instead a robust empirical spread of \(Q\) across candidate masks (Eq.~\ref{eq:mask_uncertainty}), not a derivative term. It is combined with the others under the same independence assumption.

The wind term is
 
\begin{equation}
\sigma_{Q,\mathrm{wind}}
=
|Q|
\frac{\sigma_U}{\max(U,\epsilon)},
\label{eq:wind_uncertainty}
\end{equation} 
where \(U\) is the effective wind speed and \(\sigma_U\) is its uncertainty. This term is retained explicitly because the IME-based emission-rate equation is directly proportional to wind speed, so uncertainty in \(U\) produces the same relative uncertainty in \(Q\). Wind uncertainty is an important component of methane source-rate estimation, and controlled-release studies have shown that wind selection and environmental variability can strongly influence retrieval-based emission estimates \citep{conrad2023robust}.
 
The retrieval term propagates uncertainty in the methane enhancement raster into
IME and then into the emission rate. If pixel-level enhancement uncertainty
\(\sigma_{C_i}\) is available, the first-order retrieval contribution to IME is
 
\begin{equation}
\sigma_{\mathrm{IME,ret}}
=
F
\left[
\sum_{i \in M}
(A_i \sigma_{C_i})^2
\right]^{1/2},
\label{eq:ime_retrieval_uncertainty}
\end{equation}
where \(F\) is the concentration-to-mass conversion factor of Eq.~\ref{eq:ime} (the same \(F_{\mathrm{MLS}}(z)\) applied to every mask of a given plume; Section~\ref{sec:meteorology_conversion}), \(A_i\) is the pixel area, and \(\sigma_{C_i}\) is the per-pixel enhancement uncertainty.
The corresponding rate uncertainty is
\begin{equation}
\sigma_{Q,\mathrm{ret}}
=
3600
\frac{U}{\max(L,\epsilon)}
\sigma_{\mathrm{IME,ret}}.
\label{eq:rate_retrieval_uncertainty}
\end{equation}
This term captures the effect of uncertainty in the concentration field for the selected mask, while keeping the mask boundary, wind speed, plume length, and conversion factor fixed.
The per-pixel uncertainty raster is interpreted as a 1\(\sigma\) enhancement uncertainty, and Eq.~\ref{eq:ime_retrieval_uncertainty} 
assumes that pixel-level errors are independent. Spatially correlated retrieval errors would increase this term. We treat this independence assumption as a known 
simplification and revisit it in Section~\ref{sec:discussion_limitations}.

The mask term represents the uncertainty associated with plume boundary
selection. It is estimated from the spread of recomputed emission rates across a
family of alternative segmentations of the same plume,
 
\begin{equation}
\sigma_{Q,\mathrm{mask}}
=
1.4826\,\mathrm{MAD}\!
\left(
\{Q(M_k): M_k \in \mathcal{M}^{*}\}
\right),
\qquad |\mathcal{M}^{*}| \ge 4,
\label{eq:mask_uncertainty}
\end{equation} 
where the recomputed rates \(Q(M_k)\) are taken over the candidate family \(\mathcal{M}^{*}\), and \(\mathrm{MAD}\) is their median absolute deviation about their own median, \(\mathrm{MAD}(\{x_k\}) = \operatorname{median}_{k}\lvert x_k - \operatorname{median}_{j} x_j\rvert\). The multiplier \(1.4826 = 1/\Phi^{-1}(3/4) \approx 1/0.6745\), where \(\Phi^{-1}\) is the standard-normal quantile function, rescales the MAD so that \(1.4826\,\mathrm{MAD}\) is a consistent estimator of the standard deviation for normally distributed data. This robust spread is used in place of the ordinary sample standard deviation so that a single outlier segmentation (for example, a mask that merges a neighbouring plume) cannot inflate the mask term. The candidate family
\(\mathcal{M}^{*}\) is constructed by varying the main controls of the dynamic-threshold segmentation (the source-centered crop radius and enhancement percentile) and retaining only masks that remain plausible boundary perturbations of the selected plume. This candidate family is distinct from the GA equifinal pool
\(\mathcal{P}\) defined in Section~\ref{sec:GA}.
The GA pool is conditioned on the published IME and plume length and is used for footprint-confidence diagnostics, whereas \(\mathcal{M}^{*}\) is used to estimate how much the recomputed emission rate changes under plausible alternative plume boundaries. Numerical perturbation settings and acceptance criteria are reported in Supplementary Section~S4.
The plume length term is
\begin{equation}
\sigma_{Q,\mathrm{len}}
=
|Q|
\frac{\sigma_L}{\max(L,\epsilon)}.
\label{eq:length_uncertainty}
\end{equation} 
In the reported experiments, \(\sigma_L\) is represented using a pixel-scale digitization model based on uncertainty at the two plume endpoints, \(\sigma_L = \sqrt{2}(0.5\Delta)\), where \(\Delta\) is the pixel spacing. This term captures the effect of finite pixel resolution on the length measurement.
The mask term and the length term are deliberately kept distinct: \(\sigma_{Q,\mathrm{mask}}\) (Eq.~\ref{eq:mask_uncertainty}) represents 
boundary-selection variability in the chosen plume pixels and therefore already includes the coupled effect of boundary changes on both IME and 
mask-derived length, whereas \(\sigma_{Q,\mathrm{len}}\) represents only sub-pixel endpoint digitization for a \emph{fixed} boundary. Because the 
length term is conditioned on a fixed mask, it is treated as independent of the boundary-selection term and is not double-counted.
The conversion factor term is
\begin{equation}
\sigma_{Q,\mathrm{fac}}
=
|Q|
\frac{\sigma_F}{\max(F,\epsilon)},
\label{eq:factor_uncertainty}
\end{equation}
where \(\sigma_F\) is the uncertainty in the concentration-to-mass conversion factor \(F\). In the reported experiments, \(\sigma_F\) is a systematic relative term with \(\sigma_F/F \approx 2.1\%\), computed in quadrature from a spectroscopic-representativeness contribution (\(\approx 2\%\)) and a ground-elevation contribution (a \(50\)~m orography uncertainty propagated through \(F\), \(\approx 0.5\%\)). This value is consistent with the spread among physically reasonable pressure--temperature conventions (Section~\ref{sec:results_conversion}; Supplementary Section~S4). Because this uncertainty is fully correlated across the pixels of a plume, it is applied once as a relative term on \(Q\) rather than propagated per pixel.
For interpretation, each uncertainty component is also expressed as a variance
fraction,
\begin{equation}
f_k
=
\frac{\sigma_{Q,k}^{2}}{\sigma_Q^{2}},
\label{eq:variance_fraction}
\end{equation}
where \(k\) denotes one of the uncertainty components in
Eq.~\ref{eq:total_uncertainty}. These fractions identify whether the rate
uncertainty is dominated by wind, retrieval, mask boundary, plume length, or
conversion factor assumptions. The propagated uncertainty is used for product
consistency assessment and sensitivity interpretation. It is not a substitute for
controlled release validation of the true emission rate.
\subsection{Consistency and comparison metrics}
PlumeQuant evaluates consistency with two complementary metric classes. The first class compares mask geometry against the product reference mask. The second class compares recomputed plume quantities against published plume quantities.

The validation target is product-level mutual consistency, not independent physical truth: published quantities serve as product-reference values and the distributed mask as a product-reference boundary, with none treated as ground truth for the true plume extent or source rate. Table~\ref{tab:validation_target} summarizes how each comparison should be interpreted.
\begin{table*}
\centering
\caption{What each PlumeQuant comparison tests, and what it does not. All four comparisons assess agreement among the distributed product components under the stated conventions. None of them tests the products against independent field measurements of the true plume or the true source rate.}
\label{tab:validation_target}
\small
\setlength{\tabcolsep}{5pt}
\renewcommand{\arraystretch}{1.25}
\begin{tabularx}{\linewidth}{@{}
>{\raggedright\arraybackslash}p{0.20\linewidth}
>{\raggedright\arraybackslash}p{0.24\linewidth}
>{\raggedright\arraybackslash}X
@{}}
\toprule
\textbf{Comparison} &
\textbf{Compared against} &
\textbf{What it tests (and what it does not)} \\
\midrule
Reference-mask recomputation &
Published plume quantities, uncertainty, and the distributed mask &
Whether the delivered raster, mask, and metadata reproduce the published numbers under our stated conventions. It does not test whether the published rate or boundary is physically correct. \\
\addlinespace[3pt]
CM-like mask comparison &
Distributed mask (spatial overlap) and published plume quantities (scalar agreement) &
How closely a mask built without seeing the reference mask or the published values can reproduce both. It is not Carbon Mapper's own algorithm, and close agreement does not prove that either mask is correct. \\
\addlinespace[3pt]
GA candidate masks and footprint confidence &
Published IME and plume length (used as search targets) &
Whether many different but plausible boundaries can reproduce the same published values, that is, how strongly those values pin down the plume shape. It does not measure segmentation accuracy or give the probability of the true boundary. \\
\addlinespace[3pt]
Uncertainty reconstruction &
Published emission-rate uncertainty &
Whether the published uncertainty can be rebuilt from the wind, retrieval, mask, length, and conversion terms. It is not a complete error budget and does not replace controlled-release validation. \\
\bottomrule
\end{tabularx}
\end{table*}
 
For spatial comparison, the primary mask overlap metrics are intersection over union and the Dice coefficient. For a candidate mask \(M_j\) and product reference mask \(M_{\mathrm{ref}}\), intersection over union is
 
\begin{equation}
\mathrm{IoU}(M_j,M_{\mathrm{ref}})
=
\frac{
|M_j \cap M_{\mathrm{ref}}|
}
{
|M_j \cup M_{\mathrm{ref}}|
},
\label{eq:iou}
\end{equation}
which corresponds to the Jaccard similarity index \citep{jaccard1912distribution}. The Dice coefficient is
 
\begin{equation}
\mathrm{Dice}(M_j,M_{\mathrm{ref}})
=
\frac{
2|M_j \cap M_{\mathrm{ref}}|
}
{
|M_j| + |M_{\mathrm{ref}}|
},
\label{eq:dice}
\end{equation}
which emphasizes the shared portion of the two masks \citep{dice1945measures}. These metrics summarize geometric agreement between a candidate plume boundary and the product reference boundary.
 
For plume quantity comparison, the relative difference between a recomputed quantity and the corresponding published quantity is calculated as
 
\begin{equation}
\Delta_X(M_j)
=
100
\frac{
X(M_j)-X_{\mathrm{pub}}
}
{
\max(|X_{\mathrm{pub}}|,\epsilon)
},
\label{eq:relative_difference}
\end{equation}
where \(X \in \{{\mathrm{IME}, L, Q}\}\), \(X(M_j)\) is the value recomputed from mask \(M_j\), and \(X_{\mathrm{pub}}\) is the corresponding published value. Values of \(\Delta_X(M_j)\) near zero indicate close agreement between recomputed and published quantities under the stated PlumeQuant assumptions. Larger differences can reflect differences in mask geometry, plume-length convention, wind input, concentration-to-mass conversion factor, product version, or uncertainty treatment.

When both recomputed and published emission rate uncertainties are available, rate consistency is also evaluated with an uncertainty-normalized difference,
\begin{equation}
Z_Q(M_j)
=
\frac{
|Q(M_j)-Q_{\mathrm{pub}}|
}
{
\left[
\sigma_Q(M_j)^2
+
\sigma_{Q,\mathrm{pub}}^2
\right]^{1/2}
}.
\label{eq:normalized_rate_difference}
\end{equation}
Here, \(Q(M_j)\) and \(\sigma_Q(M_j)\) are the emission rate and propagated uncertainty recomputed for mask \(M_j\), while \(Q_{\mathrm{pub}}\) and \(\sigma_{Q,\mathrm{pub}}\) are the corresponding published values. This metric compares the recomputed--published rate difference with the combined uncertainty scale. It is used as a diagnostic measure of product consistency, not as a binary validation test.
 
These metrics are interpreted jointly: a mask can reproduce the published scalars yet overlap the reference poorly, or match the reference geometry yet yield a different rate under different length, wind, or conversion assumptions. These patterns identify whether differences arise from geometry, meteorology, or convention.
\subsection{Software implementation and reproducibility}
PlumeQuant was implemented as a reproducible local analysis workflow for plume-level consistency assessment. The scientific processing backend is written in Python and reads plume metadata, geospatial rasters, meteorological inputs, and mask products; generates CM-like and GA candidate masks; recomputes IME, plume length, emission rate, uncertainty, and comparison metrics; and writes analysis exports. Core processing uses NumPy, SciPy, \texttt{rasterio}, pandas, and DEAP for the genetic algorithm. A local browser interface is used for visualization, mask review, and export management, but all reported quantities are produced by the backend computation workflow. Outputs are written with provenance information: product identifiers, raster paths, meteorological source, conversion-factor convention, mask source, random seed, and configuration settings. All deterministic computations are reproducible from the input products and configuration, and GA-based outputs are reproducible when the fixed random seed and stored configuration are used. The released repository also includes data-acquisition scripts for retrieving public Carbon Mapper plume-product files and NOAA HRRR meteorological fields from the identifiers and timestamps listed in the plume manifest. Implementation details, cache conventions, and reproducibility checks are provided in Supplementary Section~S1.

\section{Results}
\label{sec:results}
 
All results were computed for 63 publicly available Carbon Mapper methane plume records generated from EMIT observations for which source location, IME, plume length, emission rate, and the required raster products were available. 
Emission-rate and uncertainty comparisons use the 62 of these records that also carry a published emission rate and rate uncertainty
 (see Section~\ref{sec:data_products}); IME, plume-length, and mask-overlap comparisons use all 63.
When uncertainty comparisons are reported, only records with the corresponding published uncertainty metadata are included. 
When development and holdout results are reported, the split was fixed a priori at the scene level: plumes sharing a scene raster were grouped, and each scene group was assigned to the development or holdout set by a deterministic seeded hash with a 35\% holdout fraction, so the two sets are scene-disjoint, the assignment is independent of all published plume quantities, and it remains stable as records are added. This yields 44 development plumes from 18 scenes, on which all calculation conventions and mask-generation settings were selected, and 19 holdout plumes from 9 scenes, scored once after the configuration was frozen. Scene-level split assignments and per-plume results are listed in Supplementary Section~S5.
Unless otherwise stated, signed differences are reported as percentage differences relative to the published product value, \(100[ X_{\mathrm{rec}}-X_{\mathrm{pub}}]/X_{\mathrm{pub}}\), and absolute difference denotes the absolute value of this quantity. These differences are product-consistency metrics, not errors relative to controlled-release, in situ, or other independent truth measurements. 
The reference-mask recomputation primarily tests whether the inferred conversion and length conventions reproduce the published scalars. The CM-like comparison is the stronger test because its mask never sees the reference mask or any published value: it is generated only from the enhancement raster and the reported source location, with settings fixed once on the development split (Section~\ref{sec:cm_like_mask}). Unless stated otherwise, end-to-end agreement is computed from the CM-like mask.
 
\subsection{Concentration-to-mass conversion factor}
\label{sec:results_conversion}
Eight concentration-to-mass conversion conventions were evaluated against the published IME values without applying empirical scaling or calibration (Table~\ref{tab:conversion_conventions}). The comparison was designed to isolate the effect of the physical conversion convention rather than differences in software implementation. When HRRR meteorological fields were available, the same cached HRRR inputs and sampling workflow were used across the HRRR-based conventions. Ground elevation was taken from HRRR model orography. This setup allowed the conversion-factor comparison to test whether pressure--temperature assumptions could explain residual differences between recomputed and published IME.
\begin{table*}
\centering
\caption[Conversion-convention agreement with published IME]{Conversion-convention agreement with published IME on the development split (\(n=44\)).}
\label{tab:conversion_conventions}
\small
\setlength{\tabcolsep}{4pt}
\renewcommand{\arraystretch}{1.10}
 
\begin{tabularx}{\linewidth}{@{}
>{\raggedright\arraybackslash}X
>{\raggedleft\arraybackslash}p{0.15\linewidth}
>{\raggedleft\arraybackslash}p{0.15\linewidth}
>{\raggedleft\arraybackslash}p{0.11\linewidth}
@{}}
\toprule
\textbf{Convention} &
\shortstack{\textbf{Median}\\\(\boldsymbol{|\Delta|}\) \textbf{(\%)}} &
\shortstack{\textbf{Median}\\\textbf{bias (\%)}} &
\shortstack{\(\boldsymbol{r(T_{2m})}\)} \\
\midrule
Std. atmosphere at ground elevation (adopted) & \textbf{0.07} & -0.07 & 0.05 \\
HRRR PBL hybrid & 2.12 & -1.50 & -0.97 \\
Local surface, total air & 2.78 & -2.66 & -0.99 \\
Local surface, dry air & 4.16 & -4.16 & -0.98 \\
U.S. Standard Atmosphere 1976 & 2.38 & +2.38 & -0.24 \\
Local PBL mean, total air & 8.54 & -8.54 & -0.93 \\
Local PBL mean, dry air & 9.02 & -9.02 & -0.95 \\
NTP fallback & 17.76 & +17.76 & -0.43 \\
\bottomrule
\end{tabularx}
 
\vspace{2pt}
\begin{minipage}{\linewidth}
\footnotesize
\emph{Note:} Median \(|\Delta|\) is the median absolute percentage difference from the published IME.
\(r(T_{2m})\) is the Pearson correlation between the signed residual and 2~m temperature.
The adopted standard-atmosphere convention corresponds to the midlatitude-summer profile evaluated at ground elevation.
\end{minipage}
\end{table*}

\begin{figure*}
\centering
\includegraphics[width=0.9\textwidth]{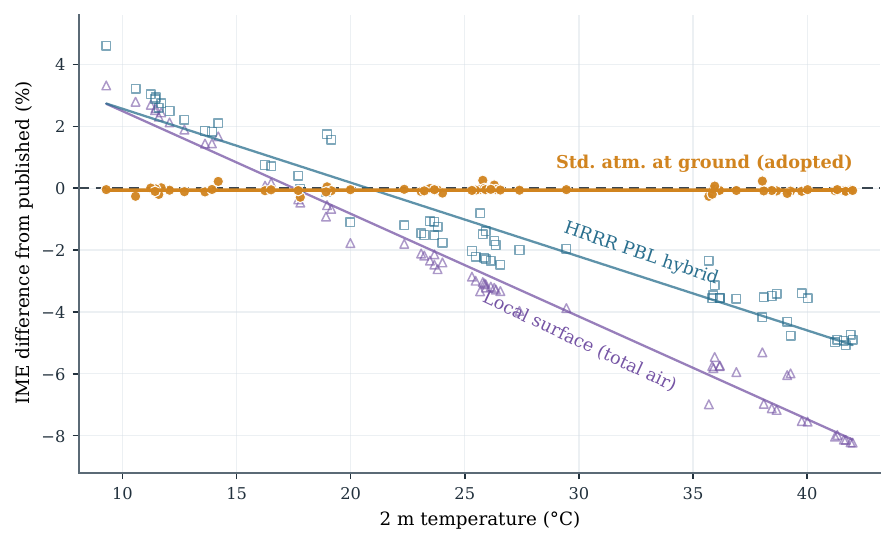}
\caption{Conversion-factor residuals relative to published IME as a function of near-surface temperature. Points show plume-level residuals, and solid lines show linear trend lines for each conversion convention. The adopted midlatitude-summer standard-atmosphere convention remains near zero across the temperature range, whereas the HRRR boundary-layer hybrid and local-surface (total-air basis; the $-2.66\%$ median-bias row of Table~\ref{tab:conversion_conventions}) conventions show strong temperature-dependent residuals.}
\label{fig:factor_residual_temperature}
\end{figure*}

The midlatitude-summer standard-atmosphere conversion convention evaluated at ground elevation provided the closest agreement with published IME and was therefore used for subsequent recomputation. On the development subset, its median absolute difference from published IME was 0.07\%, with a median signed bias of -0.07\% and a median absolute deviation of 0.03\%. On the holdout subset, the corresponding values were 0.06\%, -0.06\%, and 0.02\%. The residuals showed little dependence on near-surface temperature or boundary-layer height, with Pearson correlation coefficients of \(r = 0.05\) for 2~m temperature and approximately zero for boundary-layer height. Temperature-tercile medians were also stable, with residuals of -0.07\% for cold scenes, -0.05\% for middle-tercile scenes, and -0.07\% for warm scenes. This behavior is shown in Figure~\ref{fig:factor_residual_temperature}, where the adopted convention remains centered near zero across the temperature range, unlike the HRRR hybrid and local-surface conventions, which show stronger temperature-dependent residuals.

Alternative conversion conventions showed larger disagreement with published IME and, in several cases, introduced temperature-dependent residuals. The HRRR boundary-layer hybrid convention had a median absolute difference of 2.12\% and a strong negative correlation with 2~m temperature, \(r(T_{2m}) = -0.97\), with median residuals shifting from +1.8\% in cold scenes to -3.6\% in warm scenes. The local-surface and PBL-mean conventions were generally biased low, with median absolute differences ranging from 2.78\% to 9.02\% and temperature-correlation magnitudes of approximately 0.93--0.99. The U.S. Standard Atmosphere 1976 convention was biased high by 2.38\%, while the fixed NTP fallback was biased high by 17.76\%. Based on this comparison, the midlatitude-summer standard-atmosphere convention evaluated at ground elevation was adopted for subsequent recomputation, and the remaining conversion-factor uncertainty was represented as a systematic relative term of approximately 2.1\% in the uncertainty model.
This term is computed in quadrature from a spectroscopic-representativeness contribution ($2\%$) and a ground-elevation contribution (a $50$~m orography uncertainty
propagated through $F$, $\approx0.5\%$; Supplementary Section~S4). The resulting $\approx2.1\%$ is corroborated by the spread between the adopted convention
and the most plausible meteorologically informed alternative, the HRRR boundary-layer hybrid (median absolute difference 2.12\%; Table~\ref{tab:conversion_conventions}),
so it is a conservative estimate of how much the choice among physically reasonable pressure--temperature conventions can shift IME. It is treated as a systematic term rather than a per-plume random error. Following the GUM treatment of Type~B contributions, this systematic allowance is nevertheless combined in quadrature with the random terms in Eq.~\ref{eq:total_uncertainty} \citep{jcgm2008evaluation}.
\subsection{Plume-length convention}
\label{sec:results_length}

\begin{figure*}
\centering
\includegraphics[width=0.9\textwidth]{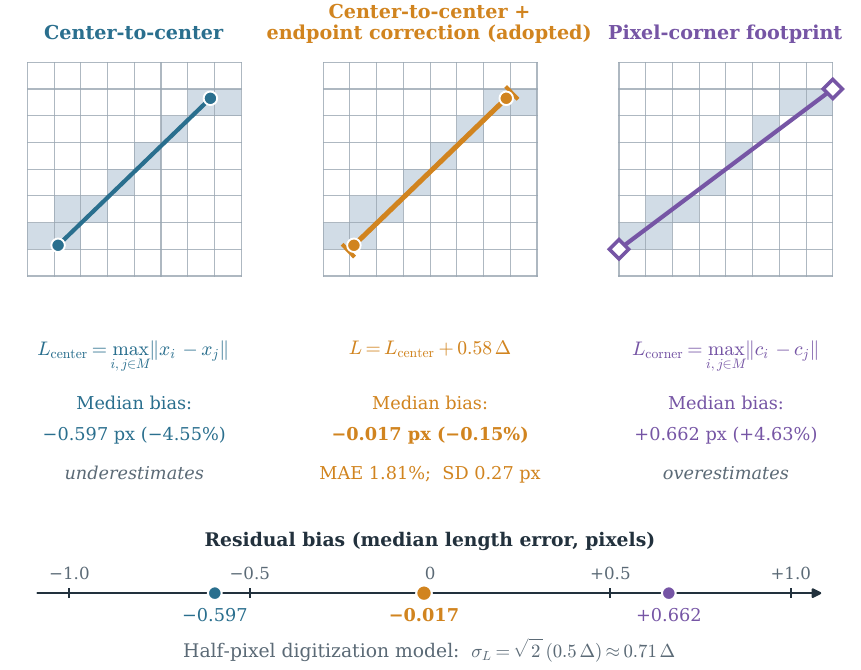}
\caption{Comparison of plume-length conventions and pixel-scale length uncertainty. The center-to-center extent underestimates published plume length, whereas the full pixel-corner footprint overestimates it. The adopted center-to-center convention with a \(0.58\,\Delta\) endpoint correction produces near-zero median bias and was used for subsequent emission-rate recomputation. The lower schematic shows the per-end half-pixel digitization model used for plume-length uncertainty, which gives \(\sigma_L = \sqrt{2}(0.5\,\Delta)\).}
\label{fig:length_conventions_schematic}
\end{figure*}

Three source-independent plume-length conventions were compared against the published plume lengths for all 63 plumes (Figure~\ref{fig:length_conventions_schematic}; Table~\ref{tab:length_conventions}). The center-to-center pairwise extent systematically underestimated plume length, with a median error of -0.597 pixels, equivalent to -4.55\%. In contrast, the full pixel-corner footprint extent overestimated plume length, with a median error of +0.662 pixels, equivalent to +4.63\%. These opposing biases indicate that the published plume-length convention lies between a strict center-to-center extent and a full corner-to-corner footprint extent.

The adopted plume-length convention, defined as the center-to-center extent plus a fixed endpoint correction of \(0.58\,\Delta\) (where \(\Delta\) is the pixel size), provided the closest agreement with published plume lengths. This convention reduced the median length bias to near zero and avoided the systematic underestimation of the center-to-center extent and the overestimation of the full pixel-corner extent. After selection, the same convention was used for all subsequent IME-to-rate recomputations. 
We emphasize that the \(0.58\,\Delta\) value is an empirical product-convention match for the EMIT-derived Carbon Mapper records analyzed here, not a physically 
universal endpoint correction; it should be re-evaluated for other sensors, ground sampling distances, or product versions before being reused.
Full per-plume error distributions for the three length conventions are provided in Supplementary Section~S6. 

\begin{table*}
\centering
\caption{Plume-length convention agreement with published plume lengths across the 63-plume benchmark. Three source-independent endpoint conventions are compared: the adopted center-chord extent plus a fixed \(0.58\,\Delta\) endpoint correction, the strict center-to-center extent, and the full pixel-corner footprint extent. Values are recomputed--published differences (signed median and mean absolute percentage error, and median and standard deviation of the pixel-scale error).}
\label{tab:length_conventions}
\small
\setlength{\tabcolsep}{3pt}
\renewcommand{\arraystretch}{1.10}
 
\begin{tabularx}{\linewidth}{@{}
>{\raggedright\arraybackslash}X
>{\raggedleft\arraybackslash}p{0.15\linewidth}
>{\raggedleft\arraybackslash}p{0.15\linewidth}
>{\raggedleft\arraybackslash}p{0.15\linewidth}
>{\raggedleft\arraybackslash}p{0.11\linewidth}
@{}}
\toprule
\textbf{Convention} &
\shortstack{\textbf{Median}\\\textbf{error (\%)}} &
\shortstack{\textbf{Mean}\\\(\boldsymbol{|\mathrm{error}|}\) \textbf{(\%)}} &
\shortstack{\textbf{Median}\\\textbf{error (px)}} &
\shortstack{\textbf{SD}\\\textbf{(px)}} \\
\midrule
Center-chord + 0.58 px (adopted) & \textbf{-0.15} & \textbf{1.81} & -0.017 & 0.27 \\
Center-to-center extent & -4.55 & 4.53 & -0.597 & 0.27 \\
Pixel-corner footprint extent & +4.63 & 5.80 & +0.662 & 0.31 \\
\bottomrule
\end{tabularx}
\end{table*}
 
\subsection{CM-like mask spatial agreement}
\label{sec:results_segmentation}
 
\begin{figure*}
\centering
\includegraphics[width=0.95\textwidth]{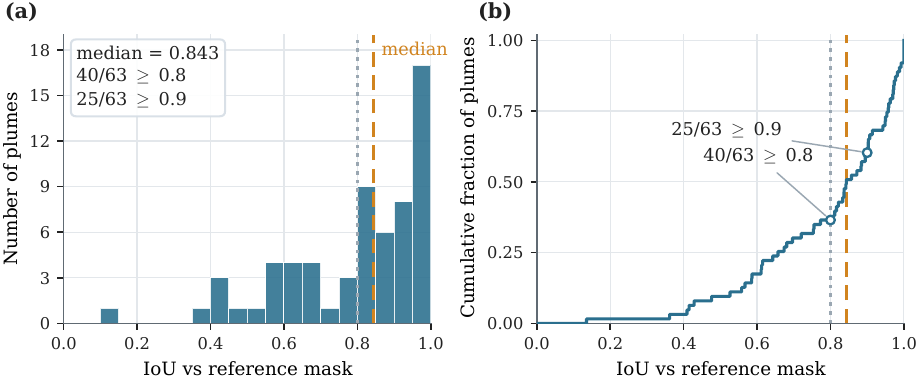}
\caption{Spatial agreement between the CM-like mask and the distributed reference mask across the 63-plume benchmark. (a) Distribution of intersection over union (IoU). The dashed line marks the median IoU and the dotted line marks IoU = 0.8. (b) Empirical cumulative distribution of the same IoU values, with the number of plumes reaching IoU $\ge$ 0.8 and $\ge$ 0.9 annotated.}
\label{fig:cmlike_iou_rate}
\end{figure*}

The CM-like mask is a non-target-conditioned baseline (Section~\ref{sec:cm_like_mask}), generated from the enhancement raster and reported source location without the reference mask or any published quantity. Constant selection showed no evidence of development-split overfitting: relative to the component-selection-only baseline, the frozen constants improved the development median IoU from 0.817 to 0.827 while leaving the holdout median IoU essentially unchanged (0.907 to 0.903). Relative to the distributed reference masks, the CM-like mask achieved a median IoU of 0.843 (95\% bootstrap CI [0.808, 0.903]; 20{,}000 resamples) across the 63-plume benchmark, with a corresponding median Dice coefficient (Eq.~\ref{eq:dice}) of 0.915 (95\% bootstrap CI [0.894, 0.949]). Forty plumes had IoU values of at least 0.8, and 25 plumes had IoU values of at least 0.9. Most plume records therefore show strong spatial agreement with the reference masks (Figure~\ref{fig:cmlike_iou_rate}).
Because the consistency metrics are computed over individual plumes while several scenes contribute more than one plume, we also 
computed a scene-clustered bootstrap that resamples whole scenes rather than individual plumes. This guards against optimistic confidence intervals arising from within-scene correlation. The scene-clustered 95\% CI for the median IoU was [0.790, 0.905], essentially unchanged from the plume-level resampling interval.

Although most CM-like masks showed strong agreement with the distributed reference masks, lower-overlap cases were concentrated in plumes with weak enhancement, fragmented plume structure, spatial offsets between the reported source and enhancement maximum, nearby artifacts, or ambiguous plume boundaries. These cases indicate that a fixed source-centered thresholding strategy can recover much of the reference plume geometry for many records, but it cannot fully resolve all plume-boundary ambiguities. Low IoU values are therefore interpreted as indicators of mask sensitivity and review priority, not as direct evidence that either mask represents an incorrect plume boundary.
 
\begin{figure*}
\centering
\includegraphics[width=0.96\textwidth]{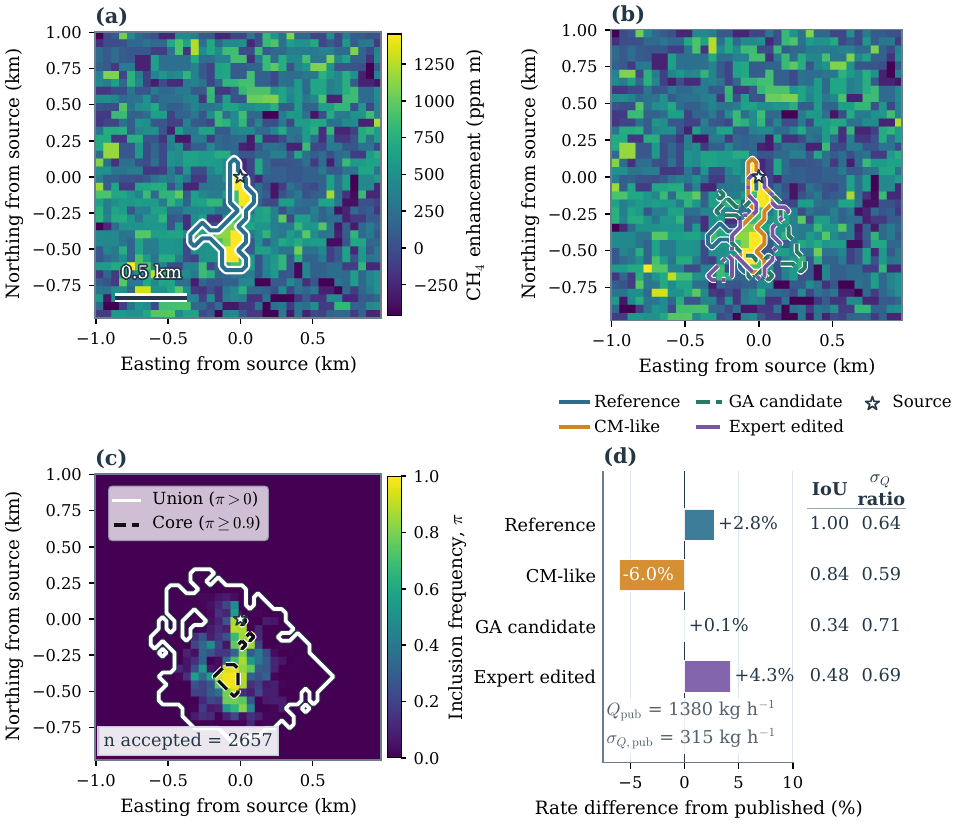}
\caption{Representative mask-sensitive plume case. 
(a) Methane enhancement raster with the distributed reference mask and reported 
source location. 
(b) Alternative mask boundaries from the reference, CM-like, GA-candidate, and 
expert-edited masks. 
(c) GA footprint-confidence map summarizing inclusion frequency across 
target-consistent masks. Values indicate mask non-uniqueness conditional on the
published IME and plume-length targets, not probability of the true plume boundary. 
(d) Recomputed emission-rate differences and uncertainty ratios across masks 
relative to the published product values. The example shows that similar 
emission-rate estimates can coexist with substantial plume-boundary sensitivity.}
\label{fig:representative_case}
\end{figure*}

Figure~\ref{fig:representative_case} shows a representative mask-sensitive plume. In this example, the reference, CM-like, GA-candidate, and expert-edited masks produce emission-rate estimates close to the published value, but their spatial overlap with the distributed reference mask differs substantially. This case illustrates a central result of the analysis: similar scalar plume quantities can be obtained from different plausible plume boundaries. Spatial agreement, scalar agreement, and uncertainty diagnostics therefore need to be interpreted jointly.
 
\subsection{Plume-boundary equifinality across the benchmark}
\label{sec:results_equifinality}
\begin{figure*}
\centering
\includegraphics[width=0.99\textwidth]{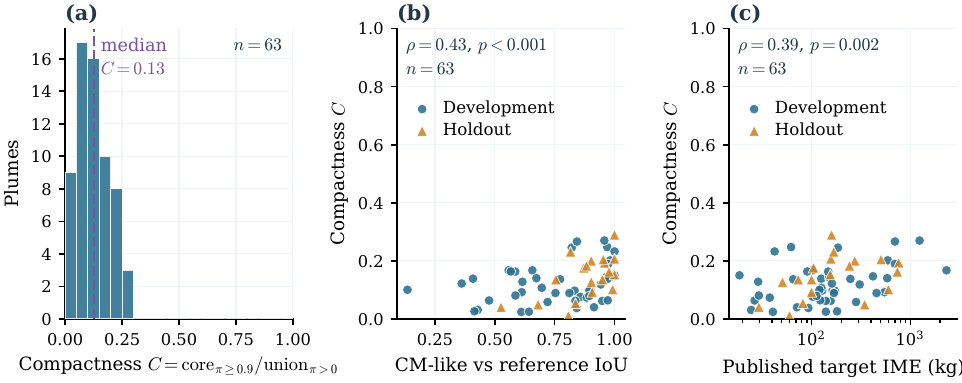}
\caption{Plume-boundary equifinality across the 63-plume benchmark. (a) Distribution of the footprint compactness ratio \(C = A_{\mathrm{core}}/A_{\mathrm{union}}\) (core: \(\pi \ge 0.9\); union: \(\pi > 0\)) over all 63 plumes, with the median marked; a low ratio indicates a large ambiguous fringe relative to the consistently selected core. (b) Compactness ratio versus CM-like IoU against the reference mask, with marker color/shape indicating the development/holdout split (Spearman \(\rho = 0.43\)). (c) Compactness ratio versus published IME (log scale; \(\rho = 0.39\)). Boundary ambiguity is pervasive and is largest for weak, low-overlap plumes. The compactness ratio is an \(R\)-conditional diagnostic (Section~\ref{sec:discussion_limitations}). The underlying inclusion-frequency maps are stable between \(R=8\) and \(R=24\).}
\label{fig:equifinality}
\end{figure*}

Figure~\ref{fig:representative_case} showed that a single plume can admit multiple target-consistent masks. To quantify this across the benchmark, we ran the GA footprint ensemble (\(R=8\) seeds, fixed GA configuration) for all 63 plumes and summarized each footprint-confidence map by a compactness ratio \(C = A_{\mathrm{core}}/A_{\mathrm{union}}\), where the core is the set of pixels selected by at least \(90\%\) of the equifinal masks (\(\pi \ge 0.9\)) and the union envelope is the set selected by any equifinal mask (\(\pi > 0\)). A high \(C\) means the published IME and plume length tightly constrain the footprint. A low \(C\) means a large ambiguous fringe surrounds a small consistently selected core. All 63 plumes yielded a reliable equifinal pool of at least 20 distinct footprints.
Across the benchmark, the consistently selected core was a small fraction of the plausible envelope: the median compactness ratio was 0.127 (IQR 0.074--0.167), so for a typical plume only about one-eighth of the target-consistent footprint was selected by nearly all equifinal masks, while the remainder formed an ambiguous fringe. The core-to-envelope gap was also large in mass terms. The IME contained in the high-confidence core and in the full union envelope differed by a median of \(76\%\) of the published IME (IQR 52--142\%). This is not disagreement among the pooled masks in reproducing the published IME, because by construction every pool member matches the published IME and plume length to within \(\pm 15\%\). Instead, it measures how loosely those scalar targets constrain the footprint: the same published IME can be spread over substantially different boundaries, so the mass attributed under a conservative (core-only) versus a generous (full-envelope) boundary differs by a large fraction of the published value.
Boundary compactness tracked the other consistency diagnostics. Plumes with more compact footprints tended to have higher CM-like IoU against the reference mask (Spearman \(\rho = 0.43\), \(p < 0.001\)) and higher published IME (\(\rho = 0.39\), \(p = 0.002\)), so the weak, low-overlap plumes evident in the CM-like IoU distribution (Figure~\ref{fig:cmlike_iou_rate}) were also the ones whose boundaries were least constrained by the published scalar quantities. Development and holdout plumes behaved similarly (median \(C = 0.103\) and 0.153). Because the union envelope can only grow as more seeds are added, \(C\) is an \(R\)-conditional diagnostic rather than an absolute quantity: increasing the ensemble from 8 to 24 seeds lowered the median \(C\) from 0.127 to 0.094, although the per-pixel inclusion-frequency maps were stable (median Spearman 0.97; mean absolute change in \(\pi\) of 0.02) and all 63 plumes retained reliable pools (Section~\ref{sec:discussion_limitations}). Figure~\ref{fig:equifinality} summarizes the compactness distribution and its relationship to spatial agreement and plume strength.
\subsection{End-to-end recomputation agreement}
\label{sec:results_end_to_end}
To evaluate the combined effect of the selected calculation choices, Table~\ref{tab:end_to_end_agreement} compares a baseline configuration with the final harmonized configuration. The harmonized configuration uses the adopted concentration-to-mass conversion convention, the adopted plume-length convention, and the fixed CM-like mask settings. Under this configuration, recomputed plume quantities from the independently generated CM-like mask showed close agreement with the published values. The median IME error improved from -1.51\% to +0.72\%, and the mean absolute IME error decreased from 24.24\% to 16.72\%. The median plume-length error decreased from +5.46\% to +0.73\%, while the mean absolute length error decreased from 28.12\% to 16.30\%. For the emission rate, the median error improved from -7.32\% to +0.16\% (95\% bootstrap CI [\(-1.99\), 1.79]), and the mean absolute error decreased from 12.55\% to 6.98\% (95\% bootstrap CI [5.47, 8.59]).  
Resampling whole scenes instead of individual plumes gave consistent intervals (scene-clustered 95\% CI: median rate error
 [\(-1.39\), 1.79]; rate MAE [5.35, 8.68]). The agreement is therefore not an artifact of multiple plumes sharing a scene. The corresponding median IME error 
 of +0.72\% had a 95\% bootstrap CI of [\(-0.08\), 3.22].
 
\begin{figure*}
\centering
\includegraphics[width=0.95\textwidth]{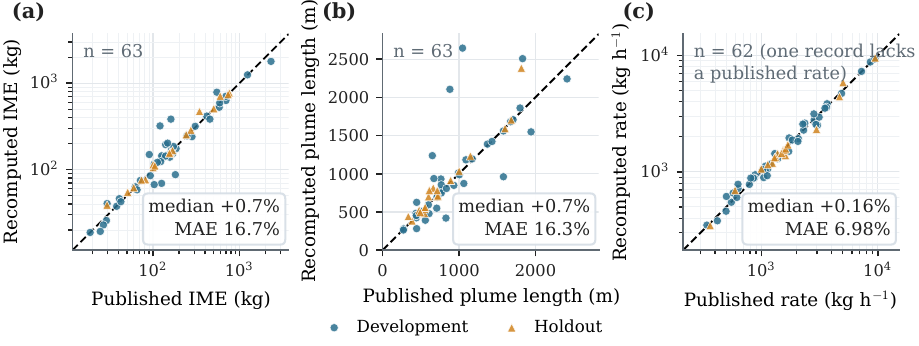}
\caption{End-to-end CM-like recomputation against published plume quantities for IME, plume length, and emission rate. Dashed lines indicate 1:1 agreement. The IME and emission-rate panels use logarithmic axes, and markers distinguish development and holdout plumes. The emission-rate panel shows the 62 records with a published rate (Section~\ref{sec:data_products}). The median and mean-absolute-error annotations in each panel are computed over the full benchmark; split-resolved statistics are reported in Table~\ref{tab:split_errors}.}
\label{fig:end_to_end_agreement}
\end{figure*}

Table~\ref{tab:split_errors} reports the same errors separately for the development split, the holdout split, and the full benchmark. Holdout performance shows no degradation: mean absolute errors on the holdout were 7.6\% for IME, 9.5\% for plume length, and 6.1\% for emission rate. The development split, on which the conventions were selected, in fact shows the \emph{larger} mean absolute errors (20.6\%, 19.2\%, and 7.3\%). This is the opposite of what tuning-induced overfitting would produce, and it has a simple cause. The mean absolute errors are dominated by a small number of weak, spatially offset, or morphologically ambiguous plumes, and the deterministic scene-level split happened to place most of these difficult plumes in the development scenes. The median errors of both splits remain close to zero. Figure~\ref{fig:end_to_end_agreement} compares recomputed and published plume quantities for all 63 plumes. Most records cluster near the 1:1 line, with the largest deviations concentrated in the difficult tail of the benchmark.

\begin{table}[t]
\centering
\caption{Harmonized-configuration CM-like errors by benchmark split: the development split (on which all conventions were selected), the scene-disjoint holdout split (scored once after the configuration was frozen), and the full benchmark. The development split carries the larger mean absolute errors, the opposite of what overfitting to the development plumes would produce, because the deterministic scene-level split placed most of the weak, spatially offset plumes in the development scenes.}
\label{tab:split_errors}
\small
\setlength{\tabcolsep}{4.5pt}
\begin{tabular}{@{}lrrrrrr@{}}
\toprule
& \multicolumn{2}{c}{Development} & \multicolumn{2}{c}{Holdout} & \multicolumn{2}{c}{All} \\
\cmidrule(lr){2-3}\cmidrule(lr){4-5}\cmidrule(l){6-7}
\textbf{Quantity} & Median & Mean $|\cdot|$ & Median & Mean $|\cdot|$ & Median & Mean $|\cdot|$ \\
\midrule
IME (\%)           & $-0.01$ & $20.65$ & $+3.22$ & $7.60$ & $+0.72$ & $16.72$ \\
Plume length (\%)  & $+0.17$ & $19.24$ & $+2.45$ & $9.51$ & $+0.73$ & $16.30$ \\
Emission rate (\%) & $-0.27$ & $7.33$  & $+0.47$ & $6.12$ & $+0.16$ & $6.98$ \\
\bottomrule
\end{tabular}

\vspace{0.4em}
\begin{minipage}{0.95\linewidth}
\footnotesize
\textit{Note.} \(n = 44/19/63\) for IME and plume length and \(n = 44/18/62\) for emission rate (one holdout record lacks a published rate).
\end{minipage}
\end{table}

The deterministic across-mask sensitivity diagnostics defined in Section~\ref{sec:uncertainty_methods} were also summarized across the benchmark. For each plume, the sensitivity range (Eq.~\ref{eq:sensitivity_range}) was computed over the available mask set (the distributed reference mask, the CM-like mask, the GA candidate mask, and the expert-edited mask when present) under the same harmonized conventions, so that the range reflects boundary choice alone. Across the 63 plumes, the median across-mask emission-rate sensitivity range \(S_Q^{\mathrm{range}}\) was 6.5\% of the reference-mask rate (IQR 3.3--11.4\%), with corresponding median ranges of 7.6\% for IME and 6.3\% for plume length. For a typical plume, plausible mask choice therefore shifts the recomputed rate by several percent, comparable to the CM-like recomputed--published rate differences and well below the median propagated rate uncertainty of approximately 31\% of the published rate, although the upper quartile shows that boundary choice moves individual records by more than 11\%.

\begin{table}[t]
\centering
\caption{End-to-end recomputation agreement with published product metadata under the pre-harmonization baseline and final harmonized configurations.}
\label{tab:end_to_end_agreement}
\small
\setlength{\tabcolsep}{5pt}
\begin{tabular}{@{}lcc@{}}
\toprule
\textbf{Quantity} &
\textbf{\shortstack{Median difference\\baseline $\rightarrow$ harmonized\\(\%)}} &
\textbf{\shortstack{Mean absolute difference\\baseline $\rightarrow$ harmonized\\(\%)}} \\
\midrule
IME, $n=63$             & $-1.51 \rightarrow +0.72$ & $24.24 \rightarrow 16.72$ \\
Plume length, $n=63$    & $+5.46 \rightarrow +0.73$ & $28.12 \rightarrow 16.30$ \\
Emission rate, $n=62$   & $-7.32 \rightarrow +0.16$ & $12.55 \rightarrow 6.98$ \\
\bottomrule
\end{tabular}

\vspace{0.5em}
\begin{minipage}{0.96\linewidth}
\footnotesize
\textit{Note.} Baseline denotes the pre-harmonization recomputation before matching the adopted concentration-to-mass conversion convention, plume-length convention, and final fixed CM-like mask settings. Harmonized denotes the final configuration using the adopted midlatitude-summer ground-elevation conversion factor, the endpoint-corrected plume-length convention \(L=L_{\mathrm{center}}+0.58\,\Delta\), and the fixed CM-like mask settings. Values are recomputed--published differences relative to published Carbon Mapper product values.
\end{minipage}
\end{table}
\subsection{Uncertainty reconstruction}
\label{sec:results_uncertainty}
The five-component first-order uncertainty envelope was reconstructed for plume records with available published uncertainty metadata and compared with the published emission-rate uncertainty (Table~\ref{tab:uncertainty_reconstruction}). Using the distributed reference mask, the reconstructed emission-rate uncertainty had a median of 443~kg~h\(^{-1}\), compared with a published median of 446~kg~h\(^{-1}\). The median reconstructed-to-published ratio for the reference mask was 1.01 (95\% bootstrap CI [0.99, 1.06]), the median signed difference was +0.7\%, and the median absolute difference was 10.3\%. Using the CM-like mask produced a similar result, with a median reconstructed uncertainty of 422~kg~h\(^{-1}\), a median ratio of 1.01, and a median absolute difference of 15.0\%.

\begin{figure*}
\centering
\includegraphics[width=0.85\textwidth]{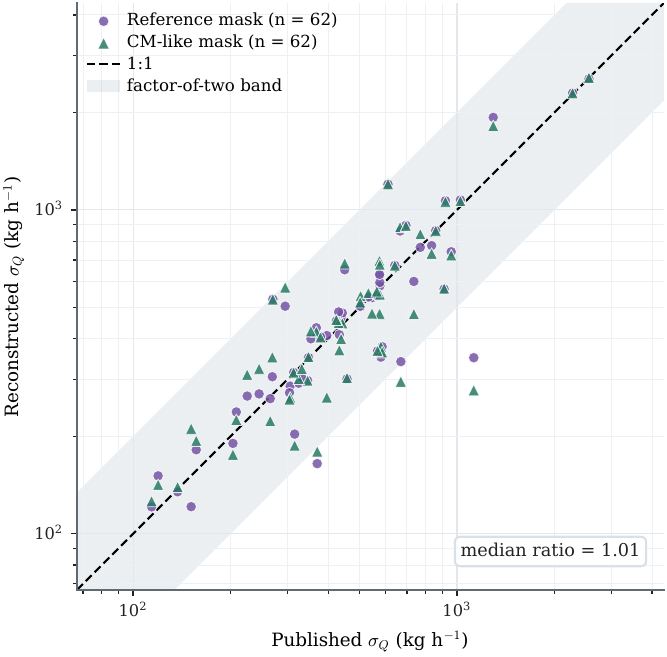}
\caption{Reconstructed versus published emission-rate uncertainty for the reference and CM-like masks. The dashed line denotes 1:1 agreement, and the shaded band denotes a factor-of-two envelope.}
\label{fig:uncertainty_reconstruction}
\end{figure*}

The mean reconstructed-to-published ratios were 1.02 for both the reference and CM-like masks, close to the corresponding medians. This indicates that a small number of high-uncertainty outliers did not dominate the comparison. The uncertainty-normalized rate differences (Eq.~\ref{eq:normalized_rate_difference}) were correspondingly small: \(Z_Q < 1\) for all 62 records with published uncertainty under both the reference and CM-like masks (median \(Z_Q\) of 0.03 and 0.09, respectively), so every recomputed--published rate difference lies well within the combined recomputed and published uncertainty scale. Overall, both mask choices reproduced the scale of the published uncertainty envelope without tuning the uncertainty model to the published uncertainty values. In Figure~\ref{fig:uncertainty_reconstruction}, most plume records fall close to the 1:1 line and remain within the factor-of-two envelope for both the reference and CM-like masks.

 Component attribution showed that wind and retrieval/IME uncertainty were the dominant contributors across the benchmark, although the dominant term varied by plume. This pattern is consistent with plume-scale wind-speed variability and with the direct dependence of the IME-based rate equation on both wind speed and retrieved excess mass. For the representative plume of Figure~\ref{fig:representative_case}, the uncertainty is dominated by the retrieval/IME term, with wind as the next largest contribution. The conversion-factor term, represented as an approximately 2.1\% relative uncertainty, was generally small in quadrature compared with the larger wind and retrieval terms, but it was retained because it represents a systematic conversion uncertainty. Mask-boundary and plume-length terms were typically secondary, although they became more important for plumes with weak enhancement, fragmented structure, or poorly constrained plume extent.
 
\begin{table*}
\centering
\caption[Reconstructed versus published emission-rate uncertainty]{Reconstructed versus published emission-rate uncertainty for plumes with published uncertainty metadata (\(n=62\)).}
\label{tab:uncertainty_reconstruction}
\small
\setlength{\tabcolsep}{3pt}
\renewcommand{\arraystretch}{1.12}
 
\begin{tabularx}{\linewidth}{@{}
>{\raggedright\arraybackslash}X
>{\raggedleft\arraybackslash}p{0.23\linewidth}
>{\raggedleft\arraybackslash}p{0.19\linewidth}
>{\raggedleft\arraybackslash}p{0.21\linewidth}
@{}}
\toprule
\textbf{Mask} &
\shortstack{\textbf{Median}\\\textbf{reconstructed}\\\(\sigma_Q\)\\\textbf{(kg h\(^{-1}\))}} &
\shortstack{\textbf{Median ratio}\\\textbf{to published}} &
\shortstack{\textbf{Median}\\\(|\mathrm{difference}|\)\\\textbf{(\%)}} \\
\midrule
Reference
& 443
& 1.01
& 10.3 \\
 
Carbon Mapper-like
& 422
& 1.01
& 15.0 \\
 
Published metadata
& 446
& ---
& --- \\
\bottomrule
\end{tabularx}
\end{table*}

\FloatBarrier
\section{Discussion}
\label{sec:discussion}
 
\subsection{Product consistency rather than independent truth}
\label{sec:discussion_consistency}
 
A natural concern about any recomputation study is circularity: conventions selected to match published values will, by construction, match them. Three design choices bound this concern. First, the CM-like mask never uses the distributed reference mask or any published quantity during mask generation. Its only inputs are the enhancement raster and the reported source location. Second, every tunable convention and constant was selected on the scene-disjoint development split and frozen before the holdout scenes were scored exactly once. Third, the equifinality analysis does not depend on agreement at all: it asks how strongly the published scalar quantities constrain the plume boundary, and its central result, a small consistently selected core inside a broad plausible envelope, would stand even if the recomputed and published quantities disagreed. With this concern bounded, the agreement results below are read strictly as consistency findings.

The central result is that most recomputation differences trace to explicit product conventions and mask choices rather than to an irreconcilable mismatch among the distributed components. PlumeQuant evaluates whether the raster, source location, reference mask, wind information, and published quantities support a common quantitative interpretation. It does not retrieve methane from radiances, detect plumes, or establish the true emission rate. Agreement metrics should be read accordingly: a small recomputed--published difference means the components can be reconciled under the stated conventions, not that the published rate is unbiased, and a large difference identifies a record whose mask geometry, wind input, conversion convention, product version, or metadata linkage warrants review.

Under matched conventions the recomputation closely reproduced the published quantities: the adopted conversion convention removed nearly all median IME bias, the adopted length convention removed the dominant pixel-scale length bias, and the reconstructed uncertainty envelope matched the published scale. Remaining disagreements concentrated in weak, spatially offset, or morphologically complex plumes rather than forming a systematic bias across the benchmark.
 
\subsection{Importance of conversion-factor and plume-length conventions}
\label{sec:discussion_conventions}
 
The conversion-factor and plume-length results show why product conventions must be matched before recomputation differences are interpreted physically. IME is directly proportional to the conversion factor, so changing the pressure--temperature convention changes IME and rate even when the mask, raster, and wind are unchanged. The midlatitude-summer convention produced the closest agreement with published IME, whereas local-surface and HRRR boundary-layer alternatives produced larger, temperature-structured residuals. This does not make the standard atmosphere a better description of each plume's boundary layer. Rather, it identifies the scalar convention most consistent with the published values, and substituting a more meteorologically local factor would introduce a methodological bias rather than a physical correction. Plume length acts similarly because the rate is inversely proportional to \(L\) in the standard IME--wind--length formulation \citep{varon2018quantifying,duren2019california,duren2025carbon}: the center-to-center extent underestimated and the pixel-corner extent overestimated the published lengths, and the adopted endpoint-corrected convention removed this bias. The plume quantities should therefore be evaluated as linked rather than independent. A rate can appear correct because IME and length errors compensate, so reporting the component differences separately is more informative than reporting only the final emission-rate difference.
 
\subsection{Mask sensitivity and the role of automated alternatives}
\label{sec:discussion_masks}
 
Plume-boundary choice affects every recomputed plume quantity even with all other inputs fixed. The CM-like mask agreed strongly with the distributed reference masks at the median, but lower-overlap cases occurred for weak, fragmented, offset, or morphologically ambiguous plumes. Such disagreement is not only a segmentation issue. It can reflect genuine ambiguity in how the boundary should be drawn from the enhancement field, so low spatial overlap is best read as a sensitivity signal and review priority rather than as evidence that either mask is wrong. The GA ensemble complements this view: because it is conditioned on the published IME and plume length, it does not provide an independent mask, but it shows whether many plausible boundaries reproduce the same scalar quantities. A compact high-confidence footprint means the published scalars strongly constrain the geometry. A broad ambiguous fringe means they do not, which is why scalar agreement alone cannot establish a unique boundary.

Expert-edited masks make analyst judgment explicit and recomputable rather than informal: passing an edited boundary through the same equations shows how much the plume quantities change under analyst-guided choices. Boundary non-uniqueness is also why spatial agreement, scalar agreement, and uncertainty must be interpreted jointly. No single metric captures where boundaries differ, whether those differences affect the reported quantities, and whether they are large relative to the uncertainty envelope.
 
\subsection{Uncertainty reconstruction and interpretation}
\label{sec:discussion_uncertainty}
The reconstructed-to-published uncertainty ratios were centered near one for both the reference and CM-like masks (Figure~\ref{fig:uncertainty_reconstruction}) without tuning the model to individual published values. This agreement indicates that the published rate uncertainties are broadly consistent with the wind, retrieval, mask-boundary, length, and conversion terms represented in PlumeQuant \citep{jcgm2008evaluation}. Wind and retrieval/IME terms dominated for most plumes, as expected from the rate equation's direct dependence on both and from controlled-release evidence that wind selection strongly influences source-rate estimates \citep{conrad2023robust}. Mask and length terms grew for weak, fragmented, or spatially ambiguous plumes. The model is a structured consistency tool, not a complete error budget: it omits retrieval-model errors, unresolved atmospheric transport, and source intermittency, so agreement supports internal consistency of the product information but does not replace controlled-release or in situ validation.
 
\subsection{Practical use of the consistency framework}
\label{sec:discussion_implications}
 
For product users, the results suggest a practical review sequence: recompute the published quantities from the distributed components under matched conventions; compare alternative masks with the reference to gauge boundary sensitivity; use the uncertainty decomposition to judge whether differences exceed the propagated envelope; and reserve expert review for plumes where weak enhancement, source offsets, nearby sources, artifacts, or ambiguous morphology leave the automated interpretation uncertain. A recomputed rate can differ from the published value through the mask, the length definition, the conversion factor, or the wind input. When these terms are treated separately, the comparison becomes transparent and shows which assumption controls the result.
 
\subsection{Limitations and future work}
\label{sec:discussion_limitations}
 
This study has two principal limitations, alongside the narrower simplifications noted below. First, the benchmark contains 63 Carbon Mapper methane plume records generated from EMIT observations. Although this is a census of the records available for the study region and period rather than a curated subset, it is drawn from a single oil-and-gas-producing region. It therefore does not represent the full range of sensors, surfaces, atmospheric conditions, source types, and plume morphologies encountered in methane remote sensing. Second, the GA result requires especially careful interpretation because it is explicitly conditioned on the published IME and plume-length targets. The CM-like mask, in turn, is a transparent, description-guided analogue and should not be interpreted as Carbon Mapper's internal operational implementation.
 
The uncertainty model is limited by the information available in the distributed products and auxiliary meteorology. It does not fully represent transport-model error, subpixel source structure, plume intermittency, retrieval biases under complex surface conditions, or unresolved vertical wind structure \citep{varon2018quantifying,foote2021impact,conrad2023robust}, and the HRRR-based wind treatment cannot resolve all local flow variability near individual sources \citep{dowell2022high,noaa_hrrr}. Future work should compare PlumeQuant outputs with controlled-release experiments, coincident observations, and larger multi-sensor datasets to separate product-consistency differences from true emission-rate errors.

Two specific simplifications should also be noted. First, the retrieval uncertainty term assumes spatially independent pixel-level errors. Correlated retrieval errors
would increase the IME and rate uncertainty, so the reported retrieval contribution is best read as a lower bound under the independence assumption. Second, 
the GA inclusion-frequency (footprint-confidence) diagnostic is computed from an ensemble of \(R=8\) seeds. We tested its sensitivity to ensemble size by re-running all 63 plumes at \(R=24\): the per-pixel inclusion-frequency maps were stable (median Spearman correlation of 0.97 between \(R=8\) and \(R=24\), and a mean absolute change in \(\pi\) of 0.02) and every plume retained a reliable pool, but the summary compactness ratio decreased with more seeds (median 0.127 at \(R=8\) versus 0.094 at \(R=24\)) because the union envelope can only grow as additional target-consistent footprints are sampled. The inclusion frequencies should therefore be read as an \(R\)-conditional, exploratory diagnostic of boundary non-uniqueness rather than as stable absolute probabilities of plume membership. The compactness summary is likewise conditional on the \(\pm 15\%\) target tolerances used to define the equifinal pool. Narrower tolerances would retain fewer and more similar footprints, and re-filtering the archived pools at tighter tolerances is a straightforward robustness extension.
 Future work should also move beyond deterministic candidate masks toward probabilistic boundary models that propagate mask uncertainty directly into the plume quantities; relate mask-ambiguity patterns to plume strength, surface type, retrieval noise, and wind regime in larger datasets; and apply the same workflow to additional sensors and product versions to test which findings generalize.
 
\section{Conclusions}
\label{sec:conclusions}
 
This study presented PlumeQuant, an uncertainty-aware consistency-assessment framework for EMIT-derived Carbon Mapper methane plume products. Recomputing the plume quantities and uncertainty from the distributed components under explicit conventions closely reproduced the published values across the 63-plume benchmark, and the reconstructed uncertainty envelope matched the published scale. The results also show that scalar agreement does not imply a unique plume boundary: different plausible masks reproduce similar plume quantities, especially for weak, fragmented, offset, or morphologically ambiguous plumes, so spatial agreement, scalar agreement, mask sensitivity, and uncertainty decomposition must be evaluated jointly.

By making calculation choices explicit and recomputable, PlumeQuant offers a practical way to review methane plume products without treating any single mask as ground truth, which grows in importance as imaging-spectrometer archives expand. Applications to larger multi-sensor datasets, controlled-release experiments, and independent field measurements can further clarify when product-consistency differences reflect calculation choices and when they indicate true emission-rate uncertainty.
 
\section*{CRediT authorship contribution statement}
 
Parisa Masnadi Khiabani: Conceptualization, Methodology, Software, Formal analysis, Visualization, Writing -- original draft. Wolfgang Jentner: Software, Visualization, Writing -- review \& editing. Alireza Rangrazjeddi: Conceptualization, Methodology, Formal analysis, Writing -- review \& editing. Michael C. Wimberly: Supervision, Writing -- review \& editing. Binbin Weng: Supervision, Writing -- review \& editing. David Ebert: Conceptualization, Supervision, Writing -- review \& editing. Charles Nicholson: Methodology, Supervision, Writing -- review \& editing.
 
\section*{Declaration of competing interest}
 
The authors declare that they have no known competing financial interests or personal relationships that could have appeared to influence the work reported in this paper.
 
\section*{Declaration of generative AI and AI-assisted technologies in the writing process}

During the preparation of this manuscript, the authors used Grammarly for grammar, spelling, punctuation, and language-editing checks, and Anthropic Claude for language editing and organizational assistance. After using these tools, the authors reviewed and edited the content as needed and take full responsibility for the content of the published article.
 
\section*{Data and code availability}

The Carbon Mapper/EMIT methane plume products analyzed in this study are publicly available through the Carbon Mapper open data platform and associated APIs. NOAA High-Resolution Rapid Refresh (HRRR) meteorological fields are publicly available through NOAA/NCEP and public NOAA cloud archives.

The PlumeQuant source code, data-acquisition scripts, configuration files, fixed random seeds, plume manifest, environment specification, scripts used to generate the manuscript tables and figures, and derived plume-level metrics are archived on Zenodo at \url{https://doi.org/10.5281/zenodo.21282534}. The original Carbon Mapper raster products and HRRR meteorological files are not redistributed. Instead, the archive provides product identifiers, acquisition scripts, cache conventions, and step-by-step instructions for retrieving the public source data.
 
\section*{Acknowledgements}
 
The authors acknowledge Carbon Mapper and the NASA EMIT team for making plume products, documentation, and instrument information publicly available. The authors also acknowledge the NOAA HRRR program for the meteorological fields used in the consistency analysis. B.W. and D.E. acknowledge partial funding support from the U.S. Department of Energy (DOE) under grant DE-FE0032292 (iM4 program). 
 
\bibliographystyle{elsarticle-harv}
\bibliography{references}
 
\end{document}